%% file: main.tex
\documentclass{article}

 \usepackage[preprint]{neurips_2026}

\usepackage[utf8]{inputenc} % allow utf-8 input
\usepackage[T1]{fontenc}    % use 8-bit T1 fonts
\usepackage{hyperref}       % hyperlinks
\usepackage{url}            % simple URL typesetting
\usepackage{booktabs}       % professional-quality tables
\usepackage{amsfonts}       % blackboard math symbols
\usepackage{nicefrac}       % compact symbols for 1/2, etc.
\usepackage{microtype}      % microtypography
\usepackage{xcolor}         % colors

\usepackage{amsmath}
\usepackage{amssymb}
\usepackage{mathtools}
\usepackage{amsthm}
\usepackage{hyperref}
\usepackage{url}
\usepackage{graphicx}
\usepackage{subcaption}
\usepackage[export]{adjustbox}
\usepackage{wrapfig}
\usepackage{booktabs}
\usepackage{algorithm}
\usepackage{algorithmic}
\usepackage{xcolor}
\usepackage{multirow}
\usepackage[dvipsnames]{xcolor}
\usepackage{thmtools}
\usepackage{thm-restate}

\theoremstyle{plain}
\newtheorem{theorem}{Theorem}[section]

\newtheorem{lemma}[theorem]{Lemma}

\theoremstyle{definition}
\newtheorem{definition}[theorem]{Definition}

\theoremstyle{remark}

\newcommand{\methodname}{GrassWalk }
\newcommand{\secondname}{GrassJump }

\title{Geometrically Principled Randomized \\Optimization for Efficient LLM Training}

\author{%
Sahar Rajabi$^{1}$,
Nayeema Nonta$^{1}$,
Sirisha Rambhatla$^{1}$ \\
\\
$^{1}$CriticalML Lab, University of Waterloo \\\\
\\
\texttt{\{srajabi, nnonta, srambhatla\}@uwaterloo.ca}
}
\begin{document}

\maketitle

\input{Sections/abstract}
\input{Sections/introduction}
\input{Sections/preleminaries}

\input{Sections/method}
\input{Sections/experiments}
\input{Sections/related_works}

\input{Sections/conclusion}

\bibliographystyle{plainnat}
\bibliography{references}

%%%%%%%%%%%%%%%%%%%%%%%%%%%%%%%%%%%%%%%%%%%%%%%%%%%%%%%%%%%%

\newpage
\appendix
\input{Sections/appendix}

%%%%%%%%%%%%%%%%%%%%%%%%%%%%%%%%%%%%%%%%%%%%%%%%%%%%%%%%%%%%
\clearpage
\newpage
\input{checklist.tex}

\end{document}

%% file: Sections/abstract.tex
\begin{abstract}
Low-rank gradient optimization for large language models is currently divided into two categories: structured methods that rigorously identify subspaces, and randomized approaches employed primarily for computational efficiency. In this work, we question the intuition behind why random projections are effective. We trace this phenomenon to the geometry of the gradient subspaces, which exhibits subspace optimization landscape has a nearly flat curvature, while a significant portion of gradient information lies outside the core subspace. Leveraging these insights, and drawing on randomized linear algebra, we theoretically establish that random low-rank projections preserve the geometry, and we introduce GrassWalk and GrassJump, algorithms that navigate the Grassmannian manifold via random walks and jumps. By coupling this randomized exploration with subspace-aware optimizer and recovering the lost gradient signals, we achieve state-of-the-art results on LLaMA-1B, LLaMA-7B, and Qwen-1.5B pretraining. Our findings reframe randomization not merely as a computational shortcut, but as a geometrically principled approach to high-dimensional optimizations.
\end{abstract}

%% file: Sections/introduction.tex
\section{Introduction}
The training of Large Language Models (LLMs) incurs substantial memory costs, largely dominated by optimizer states. While parameter-efficient methods like LoRA \citep{hu2021lora} address memory-efficiency by restricting parameters to low-rank matrices, a recent paradigm shift targets memory reduction without sacrificing full-parameter updates. Leveraging the insight that gradients evolve within a low-dimensional subspace \citep{gurari2018gradientdescenthappenstiny, schneider2024identifyingpolicygradientsubspaces}, GaLore \citep{zhao2024galorememoryefficientllmtraining} projects gradients into low-rank manifolds to compress optimizer states. This technique has catalyzed a new family of memory-efficient algorithms \citep{robert2025ldadam, rajabi2025subtrack, chen2025fira, zhu2025apollosgdlikememoryadamwlevel} that bridge the gap between resource constraints and full-rank training. Moreover, recent findings suggest that low-rank gradient methods can outperform LoRA-based techniques, as their training landscapes contain fewer spurious local minima \citep{liu2025on}. 

Low-rank gradient techniques generally fall into two categories based on employing structured subspace adjustments or random projections. Structured methods aim to identify the precise gradient subspace via techniques like singular value decomposition (SVD) \citep{zhao2024galorememoryefficientllmtraining} and similar works \citep{chen2025fira}, or PowerSGD \citep{robert2025ldadam} and subspace tracking \citep{rajabi2025subtrack, liang2024memoryefficient}. In contrast, other approaches rely on random projections, primarily to avoid the computational cost of structured methods \citep{zhu2025apollosgdlikememoryadamwlevel, zmushko2025frugal, liu2025on, chen2025a, zhao2025separate, mulrooney2025memoryefficientdifferentiallyprivatetraining, he2025subspace}.

Recently, random projections have emerged as efficient alternatives to SVD \citep{chen2025a, liu2025on, zmushko2025frugal, zhu2025apollosgdlikememoryadamwlevel}, demonstrating they can effectively replace expensive operations. However, while they match or exceed some strong structured baselines \citep{zhao2024galorememoryefficientllmtraining, chen2025fira}, a gap remains, as they still lag behind state-of-the-art optimizers.

% Recently, random projections have emerged as efficient alternatives to SVD \citep{chen2025a, liu2025on, zmushko2025frugal, zhu2025apollosgdlikememoryadamwlevel}, demonstrating they can effectively replace expensive operations. However, while they match or exceed some strong structured baselines \citep{zhao2024galorememoryefficientllmtraining, chen2025fira}, a gap remains, as they still lag behind state-of-the-art optimizers such as SubTrack++ \citep{rajabi2025subtrack} and LDAdam \citep{robert2025ldadam}. 

To date, randomization has been employed primarily for computational efficiency \citep{zhu2025apollosgdlikememoryadamwlevel, zmushko2025frugal, liu2025on, chen2025a}; however, a fundamental understanding of its efficacy and the practical impact of its inherent approximation errors remains lacking. In this work, we investigate the foundations of randomized methods from two key perspectives. First, drawing on the literature of randomized linear algebra and high dimension geometry \citep{Vershynin_2018, Woodruff_2014, halko2010findingstructurerandomnessprobabilistic}, we establish that random low-rank projections preserve the geometry of the original space within a bounded error. To determine whether this error is practically significant enough to necessitate structured subspace adjustments, we empirically analyze the optimization landscape of low-rank gradient subspaces. We suggest that this landscape is nearly flat, and the estimation error spreads evenly; thus, most subspaces perform comparably and optimization is not sensitive to the choice of subspace. Second, we isolate and address the inherent optimization bottlenecks of low-rank gradient methods \citep{chen2025fira, robert2025ldadam, zmushko2025frugal, zhu2025apollosgdlikememoryadamwlevel, rajabi2025subtrack}, investigating whether when these challenges are properly resolved, simple randomized approaches can achieve state-of-the-art performance.

Interestingly, we show that in same settings, the specific mechanism of subspace adjustment appears irrelevant: whether one employs structured updates, executes jumps between random projections, or even performs a random walk along the subspace manifold, the performance will be remarkably similar. Building on these insights, we propose two randomized low-rank gradient methods, \methodname and \secondname, which achieve state-of-the-art performance. These approaches match or surpass the strongest baselines, including structured methods, on LLaMA-1B, LLaMA-7B, and Qwen-1.5B pretraining. Specifically, \methodname and \secondname utilize random walks and random jumps on the Grassmannian manifold to update the projection subspace. They adapt the underlying optimizer to the subspace shifts and leverage the output of the underlying low-rank optimizer to restore information lost during projection. We further provide theoretical justification demonstrating why this optimizer output effectively recovers the lost signal, despite the reliance on low-rank random projections.

%% file: Sections/preleminaries.tex
\section{Diagnosing Low-Rank Gradient Optimization}
\label{sec:challenges}
Low-rank gradient methods address the memory bottleneck of LLMs by exploiting the observation that gradients during training often evolve in a low-dimensional subspace of the full parameter space \citep{gurari2018gradientdescenthappenstiny, schneider2024identifyingpolicygradientsubspaces, zhao2024galorememoryefficientllmtraining}. They reduce memory usage while supporting full-parameter updates by projecting the gradients \(G_t \in \mathbb{R}^{m \times n}\) into a subspace of dimension \(r \ll m, n\) \citep{zhao2024galorememoryefficientllmtraining, robert2025ldadam, zhu2025apollosgdlikememoryadamwlevel, chen2025fira, rajabi2025subtrack}, denoted in \eqref{eq:projection}:
\begin{equation}
\small
\label{eq:projection}
    \widetilde{G}_t = S_t^\top G_t
\end{equation}
 where \(S_t \in \mathbb{R}^{m \times r}\) denote the orthonormal basis spanning this subspace, and \(m \le n\) without loss of generality. By applying the optimizer directly within this reduced space, we obtain its output as a low-rank update $\widetilde{G}_t^O$ which is projected back to the original space via $S_t\widetilde{G}_t^O$ to complete the full weight update. 
Assuming the gradients lie within an underlying low-rank subspace of known rank \(r\), a natural approach is to construct a rank-\(r\) approximation via SVD using the top-$r$ left singular vectors \citep{zhao2024galorememoryefficientllmtraining}.
% \begin{equation}
% \small
%     G_t=U_t\Sigma_t^GV_t^\top\approx \sum_{i=1}^r \sigma_{i,t}^G u_t^i {v_t^i}^\top,\quad S_t=[u_t^1,\dots,u_t^r].    
% \label{eq:init-svd}
% \end{equation}

Although prior works suggest the existence of a low-rank core subspace in the gradient space, they also show that this subspace is not always stable and that its variations must be captured \citep{zhao2024galorememoryefficientllmtraining}. Several strategies have been developed for updating the low-rank subspace; methods such as GaLore \citep{zhao2024galorememoryefficientllmtraining} and FiRA \citep{chen2025fira} periodically compute the SVD of the gradient matrix to identify the dominant directions, however, SVD is computationally expensive and sensitive to noise \citep{rajabi2025subtrack, zhu2025apollosgdlikememoryadamwlevel, robert2025ldadam}, particularly in later training stages when the gradients are small, leading to instability and slower convergence \citep{he2025subspace}.
To mitigate these drawbacks, methods such as APOLLO \citep{zhu2025apollosgdlikememoryadamwlevel}, FRUGAL \citep{zmushko2025frugal}, GaRare \citep{liu2025on}, and RSO \citep{chen2025a} leverage random projections to reduce computational cost. On the other hand, LDAdam \citep{robert2025ldadam}, Online Subspace Descent \citep{liang2024memoryefficient}, and SubTrack++ \citep{rajabi2025subtrack}, employ iterative approximation techniques or subspace tracking to estimate dominant gradient subspaces.
\subsection{Foundation of Random Subspaces}
\label{sec:flatness}
As discussed, random projections are emerging as efficient alternatives to expensive SVD computations \citep{zhu2025apollosgdlikememoryadamwlevel, zmushko2025frugal, liu2025on, chen2025a}, achieving on-par performance with their similar structured methods. However, the intuition driving this success remains underexplored. Specifically, \textbf{what role does the core subspace play, and how can randomized methods match structured approaches without explicitly capturing it?} Drawing on randomized linear algebra, we first show that applying random projections to compress a space preserves its geometry with high probability and a rank-dependent bounded error. Next, we investigate the landscape of these subspaces to identify the optimal projection subspace. While prior works have studied the gradient space \citep{gurari2018gradientdescenthappenstiny, zhao2024galorememoryefficientllmtraining, song2025does}, to the best of our knowledge, we are the first to analyze the {\bf subspace optimization landscape}. 
\input{Theories/geometry_preservation}

Theorem 4.1 guarantees that the magnitude of individual gradient updates is predictably maintained, while Theorem 4.2 ensures that the spatial relationships and pairwise distances between gradients are not severely distorted. These results demonstrate that by applying a deterministic scaling factor, which can be absorbed into the learning rate, randomly projected gradients act as approximate isometries. Proof of these theorems are provided in Appendix \ref{app:geometry-proof}.

To further investigate the importance of finding the optimal gradient projection, we analyze the landscape on which the subspace is optimized. Formally, each subspace corresponds to a point on the Grassmannian $Gr(m, r)$, represented by an orthonormal matrix, spanning the subspace. We leverage this underlying geometry to better understand the {\bf subspace optimization} problem.  

For our analyses, we pre-trained a LLaMA-1B architecture on the C4 dataset with $r=512$. The model consists of 24 decoder layers, each containing an attention block followed by an MLP block. 
\input{Images/tangent_singular_values_3d/tex}

The {\bf subspace estimation loss} is defined in \eqref{eq:loss-function}, which measures the loss in estimating the gradient using projection matrix $S_t$.
\begin{equation}
\label{eq:loss-function}
\small
	F(S_t) = \min_{A} \|S_t A - G_t\|^2_F,
\end{equation}
At each subspace update step, we compute the derivative of this loss with respect to $S_t$ \eqref{eq:derivative}. This derivative specifies the update direction to minimize the subspace estimation error.
\begin{equation}
\small
\frac{\partial F}{\partial S_t} = 2(S_t A - G_t) A^{\top}
\label{eq:derivative}
\end{equation}
At iteration $t$ and layer $\ell$, we extract the singular values of this derivative. We report $\max \sigma_{i, t}^\ell$, {\bf the maximum \(i\)-th singular value across all layers $\ell$ within each layer type $L$} (e.g., the largest value among the 3rd singular values of all query projection layers); thus, providing an {\bf upper-bound distribution}. Figure \ref{fig:singular_value_distribution} illustrates how this distribution evolves during training across various layer types. Notably, even in the MLP down-projection (Figure \ref{fig:singular_value_distribution}-g), which exhibits the largest singular values, the range remains small and decays rapidly. In all other layer types, the leading singular values are already exceedingly small. This suggests that the error spreads across all directions, indicating a {\bf flat subspace optimization landscape} right from the early stages of training. 

These analyses clarify why random projections are effective and capable of outperforming structured subspace adjustment methods. As Figure \ref{fig:singular_value_distribution} indicates, the gradient subspace is characterized by a nearly flat curvature. In such a topology, the variance is distributed relatively evenly across many dimensions, implying that almost any randomly selected subspace will perform similar to a precisely computed one. Furthermore, the suggested flatness of the landscape can provide a distinct optimization advantage for randomized methods: the stochastic selection of subspaces naturally injects noise into the optimization trajectory, promoting broader exploration of the loss landscape that can help the optimizer escape sharp, suboptimal local minima.

While the theoretical analysis on Theorem \ref{th:norm-preservation} and \ref{th:distance-preservations} establish worst-case guarantees, a flat singular value spectrum implies that the optimization is not sensitive to the choice of subspace. In such scenarios, this geometric preservation, combined with the exploratory noise inherent to random sampling, suggest an interesting opportunity where randomized strategies are not merely computationally efficient, but may ultimately achieve better generalization compared to structured methods.

This observation, however, raises an important question: if the subspace estimation error distributes uniformly across all directions, meaning nearly any subspace should perform equally well, why do randomized methods \citep{zhu2025apollosgdlikememoryadamwlevel, zmushko2025frugal} continue to lag behind state-of-the-art structured approaches \citep{rajabi2025subtrack, robert2025ldadam}? Does the inherent bounded error (\ref{th:distance-preservations}) fundamentally widen the gap between randomized and structured methods? Or is it possible that, under the right conditions, randomized methods can match state-of-the-art performance?

\subsection{Inherent Challenges of Low-Rank Gradient Methods}
Regardless of underlying subspace adjustment methods, low-rank gradient projection comes with two primary challenges that existing methods attempt to resolve:

{\bf 1) Momentum states misalign under changing subspaces.} Stateful optimizers like AdamW derive their accelerated convergence from historical gradient statistics; specifically, moving averages of the first and second moments, accumulated within a fixed coordinate system. However, dynamically updating the projection subspace inherently rotates this coordinate framework. This induces a severe misalignment: the historical momentum becomes invalid in the current subspace, corrupting the accumulated history that Adam relies on to precondition updates. Several methods have attempted to patch this discontinuity. COAP \citep{xiao2025coapmemoryefficienttrainingcorrelationaware} tries to align subspace updates with the first momentum direction, while FRUGAL \citep{zmushko2025frugal} either resets the momenta entirely, or linearly projects old states into the new subspace. Yet, applying simple linear projections to Adam's states is mathematically flawed due to the optimizer's non-linear variance tracking. LDAdam \citep{robert2025ldadam} and SubTrack++ \citep{rajabi2025subtrack} attempt to explicitly account for this non-linearity by introducing a complex alternative formulation that treats Adam’s state as a statistical estimator.

{\bf 2) Low-rank projections sacrifice gradient information.} A low-rank projection inherently discards gradient components orthogonal to the chosen subspace, sacrificing potentially critical training signals. Several recent methods have introduced mechanisms to mitigate this information loss. For instance, LDAdam \citep{robert2025ldadam} employs an error-feedback mechanism to re-inject discarded gradients, while FRUGAL \citep{zmushko2025frugal} relies on state-free optimizers to handle them. Building on scaling insights, methods like FiRA \citep{chen2025fira} and SubTrack++ \citep{rajabi2025subtrack} apply stateful optimizer adjustments to rescale residual components. 

\input{Images/gradient_energy/tex}
Crucially, these works treat the discarded gradients as an abstract problem to patch, without quantifying the actual magnitude of the lost energy. To determine whether the orthogonal space contains a negligible tail of noise or a substantial fraction of critical gradient energy, we introduce the retention ratio $R_t$. This metric isolates the fraction of energy preserved by the low-rank approximation by dividing the Frobenius norm of the low-rank gradient by the Frobenius norm of the full-rank gradient:
\begin{equation}
\small
    R_t \;=\; \frac{\|\widetilde{G}_t\|_F}{\|G_t\|_F}
    \label{eq:fraction}
\end{equation}
For a granular perspective, we clustered the results by layer type across all decoder layers (Figure \ref{fig:energy_fraction}). While the core subspace initially captures over 50\% of the gradient energy across all clusters, this fraction drops rapidly during the early stages of training. This decline is most pronounced in deeper layers, where the fraction is consistently lower, indicating that the gradient energy in these layers is far less concentrated.  
To verify that these findings are not merely artifacts of a small rank limit, we extended our analysis to subspace ranks of 256 and 1024 (see Appendix \ref{app:energy_frac}). As expected, increasing the rank improves energy retention. Yet, even at a rank of 1024, an exceptionally high capacity for a 1B parameter model, the retained energy plateaus at only 70–80\%. This confirms that a substantial 20–30\% of the gradient energy remains persistently scattered in the orthogonal space. At rank of 256, the captured energy drops further to 40–50\%. This robustly validates the trends in Figure \ref{fig:energy_fraction}, proving that the discarded orthogonal energy is not an edge case, but a significant structural factor that low-rank training methods must account for that.

Having analyzed the flat curvature of subspace optimization landscape and the inherent challenges of low-rank optimization, we return to our central question: is structured estimation of the core subspace truly necessary for optimal performance? Precisely, if randomization preserves the geometry and flat curvature renders most subspaces comparable, can we achieve state-of-the-art results simply by mitigating the inherent challenges of low-rank gradient methods? To answer this, we conduct a systematic ablation study. By evaluating combinations of subspace adjustment techniques while explicitly addressing these optimization challenges, we isolate the true drivers of model performance. 

\subsection{Systematic Ablation for Deconstructing Low-Rank Optimization}
To compare the effectiveness of different subspace update techniques, we consider four update methods in this ablation: \textbf{(a) Grassmannian subspace tracking} from SubTrack++ \citep{rajabi2025subtrack} that tracks the subspace by minimizing a projection error and updating along a Grassmannian geodesic (see their paper for more details). \textbf{(b) Random walk on Grassmannian} that starts from the low-rank SVD of the first gradient matrix, and then takes random walks on the Grassmannian manifold to explore the space; \textbf{(c) Random projections}, which recompute a fresh orthonormal basis at each subspace update step; and \textbf{(d) SVD-based updates} as in GaLore \citep{zhao2024galorememoryefficientllmtraining}.

We must control for the two fundamental challenges discussed previously. To isolate these variables, we integrate two mitigating components into our ablation. First, to prevent momentum state misalignment, we utilize an Adaptive Optimizer (AO) that rotates Adam's moments onto the new basis upon every subspace update, ensuring continuous alignment as seen in SubTrack++ \citep{rajabi2025subtrack} and LDAdam \citep{robert2025ldadam}. Second, to address discarded gradient information, we leverage Recovery Scaling (RS) to compensate for the lost energy, mirroring techniques used in FiRA \citep{chen2025fira}, SubTrack++ \citep{rajabi2025subtrack}, and APOLLO \citep{zhu2025apollosgdlikememoryadamwlevel}, for which we will further provide theoretical analysis, showing its effectiveness along randomized methods. We incorporate AO and RS both individually and jointly into each subspace update method, and compare the resulting evaluation losses. Further implementation details regarding the AO and RS components are provided in Section \ref{sec:method} and Appendix \ref{app:ao-rs}.
\input{Images/ablation/tex} 

As shown in Figure \ref{fig:ablation}, when neither AO nor RS is applied, Grassmannian subspace tracking achieves the lowest loss among the update rules. However, after incorporating these components, randomized methods achieve on-par results, and even slightly outperform that.
We attribute the comparable performance of randomized methods and subspace tracking to how the latter navigates the subspace landscape. Although SubTrack++ optimizes the subspace estimation loss to determine the update direction, it restricts its updates to a rank-1 alteration of the previous subspace; specifically, adjusting the direction that carries the most error. This constrained, history-aware update prevents the method from being overly influenced by gradient noise. SVD, on the other hand, is highly susceptible to noise; it ignores previous subspace information and jumps directly to a new subspace defined entirely by a single gradient observation.

Adding AO to subspace adjustment strategies yields larger improvement compared to adding RS in nearly all settings, with the notable exception of {\bf random projections}. We attribute this to the extent to which each method tries to capture most informative directions. Grassmannian updates modify the previously learned subspace through controlled rank-1 rotations; whether it is subspace tracing or a random walk on the manifold, the divergence remains small and controlled. Similarly, SVD explicitly captures dominant directions; although it is susceptible to noise, the resulting projection still retains most of the informative components. In contrast, random projections select arbitrary subspaces that may discard salient signal. Consequently, RS plays a more critical role in this setting, as discarded information is more likely to be essential, making its recovery significantly more beneficial.

This analysis demonstrates that when lost components are properly recovered (+RS and +AO+RS settings), randomized methods match or outperform state-of-the-art structured subspace estimation. Before incorporating recovery mechanism, the performance gap stems from how effectively each method tries to utilize available information. However, as shown in Section \ref{sec:flatness}, random projections preserve the underlying geometry sufficiently well. Consequently, regardless of the energy captured in a given iteration, this geometric preservation allows us to reintroduce the lost signals back into the training process, as we will formalize in Theorem \ref{th:bounded}, ultimately achieving state-of-the-art results.

%% file: Theories/geometry_preservation.tex
\begin{definition}\label{def:gr}
    The Grassmannian \(Gr(m, r)\) is the space of all \(r\)-dimensional subspaces of \(\mathbb{R}^m\).
 \end{definition}
\begin{restatable}[\bf Norm Preservation]{theorem}{normpreservation}
\label{th:norm-preservation}
Let $S$ be a projection from $\mathbb{R}^m$ onto a random $r$-dimensional subspace uniformly distributed in $Gr(m, r)$. Let $x \in \mathbb{R}^m$ be a (fixed) point and $\varepsilon > 0$. Then with probability at least $1 - 2\exp(-c\varepsilon^2 r)$, we have:
\begin{equation}
\small
(1 - \varepsilon)\sqrt{\frac{r}{m}} \|x\|_2 \le \|S^\top x\|_2 \le (1 + \varepsilon)\sqrt{\frac{r}{m}} \|x\|_2.
\end{equation}
% For any fixed vector \( x \in \mathbb{R}^n \) and random matrix \( \mathbf{S} \in \mathbb{R}^{m \times r} \) where \( S_{ij} \sim \mathcal{N}(0, 1/r) \) i.i.d., the following holds with high probability:
% \begin{equation}
% \Pr[(1-\epsilon)\|x\|^2 \leq \|\mathbf{S^\top}x\|^2 \leq (1+\epsilon)\|x\|^2] \geq 1 - 2\exp\left(-\frac{r\epsilon^2}{8}\right).
% \end{equation}
\end{restatable}
\begin{restatable}[\bf Relative Distance Preservation]{theorem}{distancepreservation}
\label{th:distance-preservations}
    Let $X$ be a set of n vectors denoted by \( x \in \mathbb{R}^m \). Consider a projection matrix $S$ from the original space to a random $r$-dimensional subspace. Assume that \(
    r \ge (C/\varepsilon^2) \log n.
\)
 Then, with probability at least $1 - 2\exp(-c\varepsilon^2 r/2)$, the scaled projection
$S' := \sqrt{{m}/{r}} S$
is an approximate isometry on $X$:
\begin{equation}
\small
    (1 - \varepsilon)\|x - y\|_2 \le \|S'^\top x - S'^\top y\|_2 \le (1 + \varepsilon)\|x - y\|_2 \quad \text{for all } x, y \in X.
\end{equation}
\end{restatable}

%% file: Images/tangent_singular_values_3d/tex.tex
\begin{figure*}[t]
 % First Rown
 \small
    \centering
    \begin{subfigure}{0.24\textwidth}
        \centering
        \includegraphics[width=\linewidth, trim=25 25 20 40, clip]{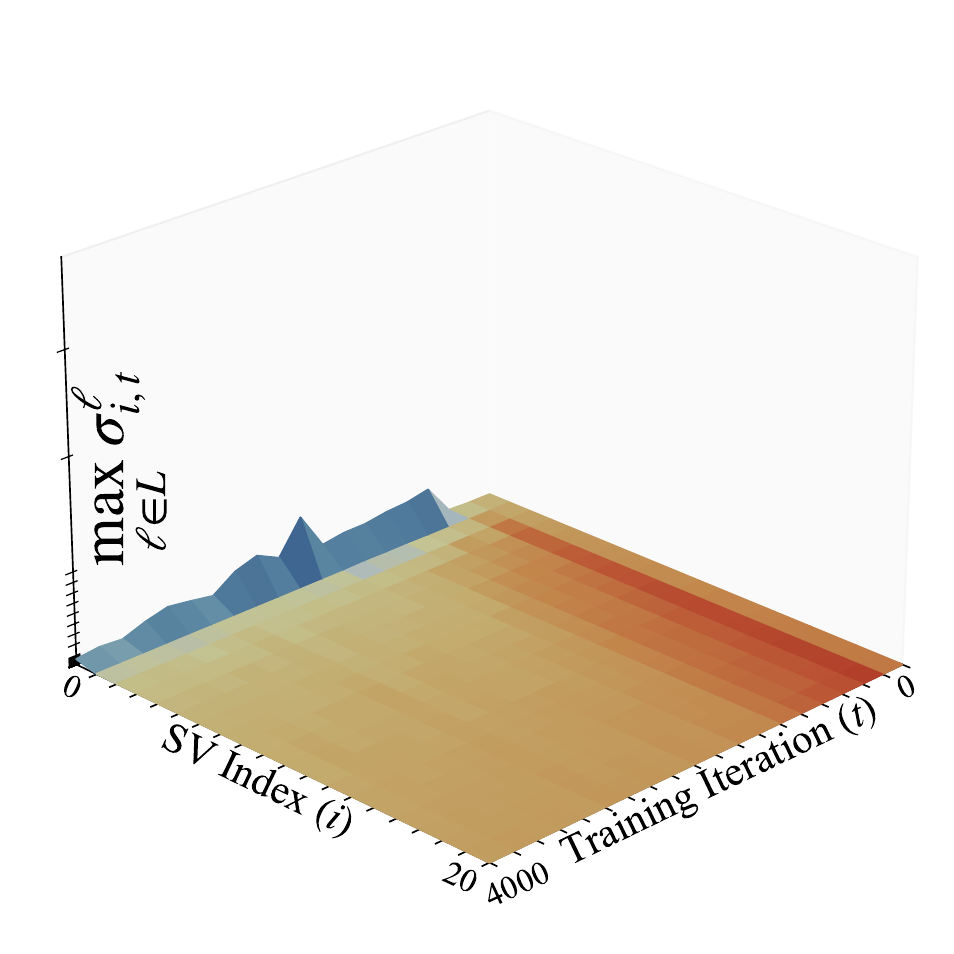}
        \caption{\scriptsize Attention-Output Proj.}
    \end{subfigure}
    \begin{subfigure}{0.24\textwidth}
        \centering
        \includegraphics[width=\linewidth, trim=25 25 20 40, clip]{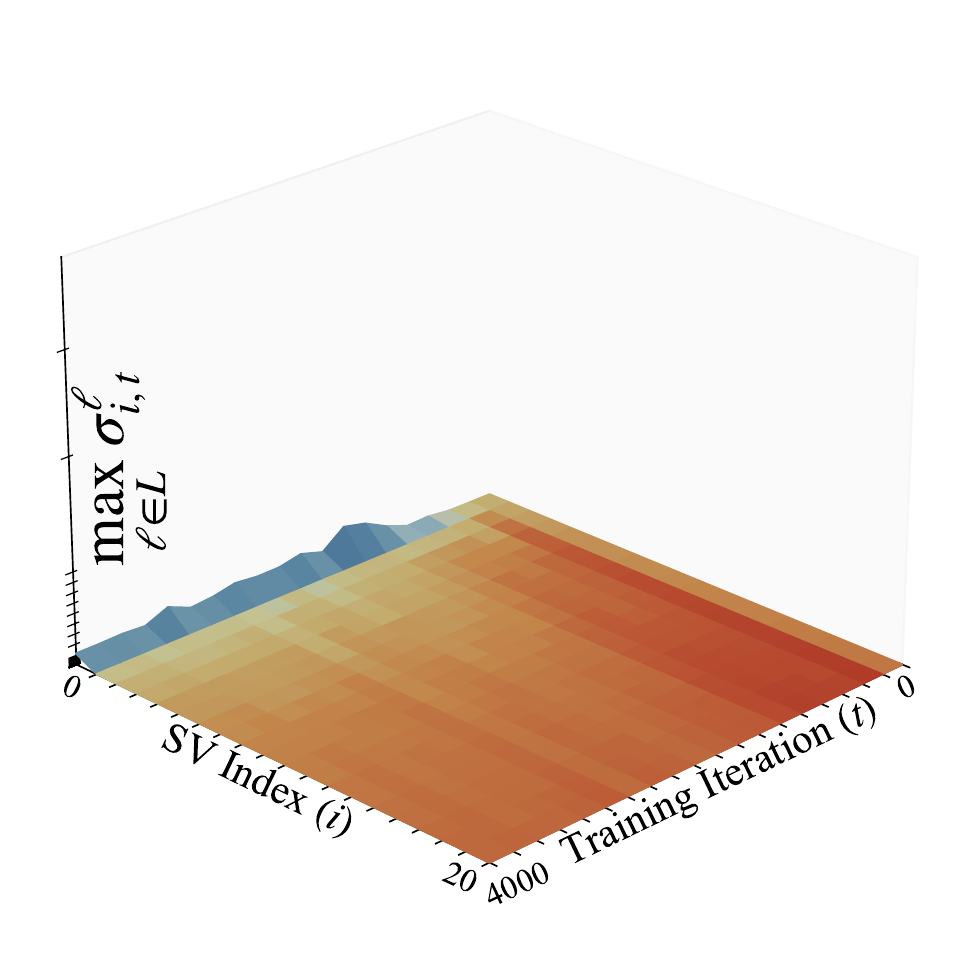}
        \caption{\scriptsize Attention-Value Proj.}
    \end{subfigure}
    \begin{subfigure}{0.24\textwidth}
        \centering
        \includegraphics[width=\linewidth, trim=25 25 20 40, clip]{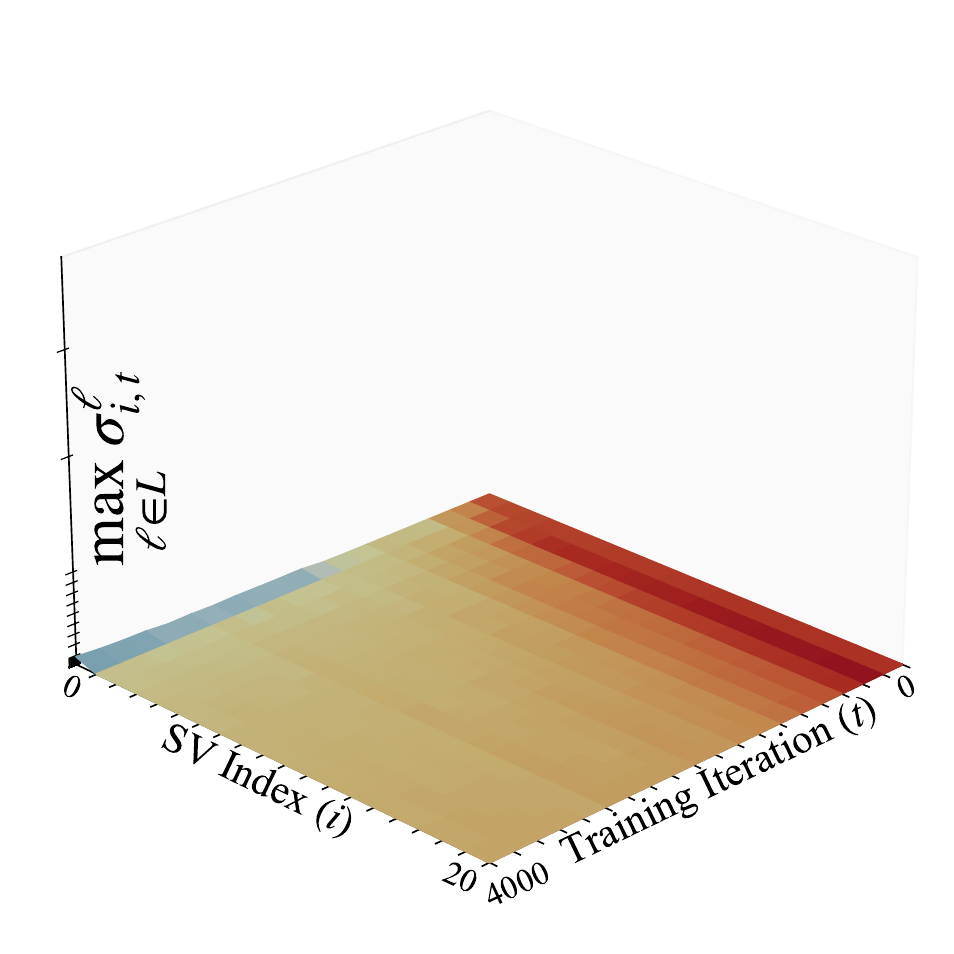}
        \caption{\scriptsize Attention-Query Proj.}
    \end{subfigure}
    \begin{subfigure}{0.24\textwidth}
        \centering
        \includegraphics[width=\linewidth, trim=25 25 20 40, clip]{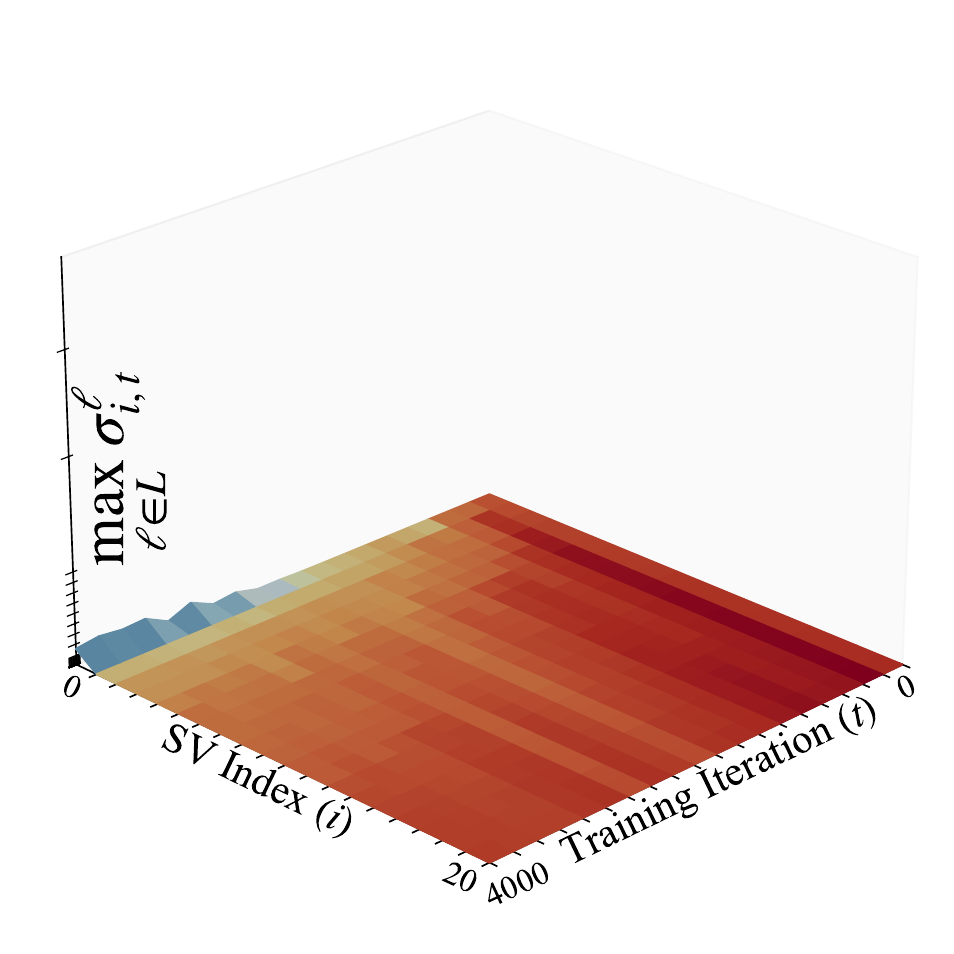}
        \caption{\scriptsize Attention-Key Proj.}
    \end{subfigure}

    % Second row
    \begin{subfigure}{0.24\textwidth}
        \centering
        \includegraphics[width=\linewidth, trim=25 25 20 40, clip]{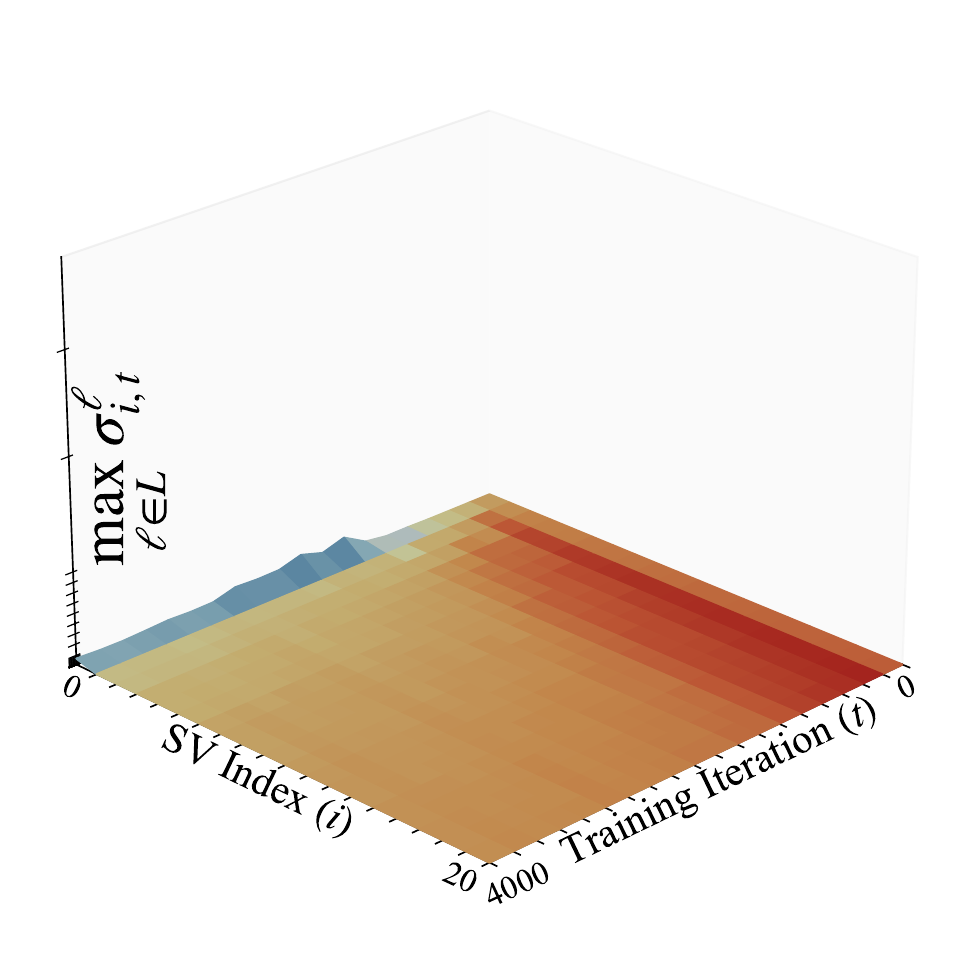}
        \caption{\scriptsize MLP-Gate Proj.}
    \end{subfigure}
    \begin{subfigure}{0.24\textwidth}
        \centering
        \includegraphics[width=\linewidth, trim=25 25 20 40, clip]{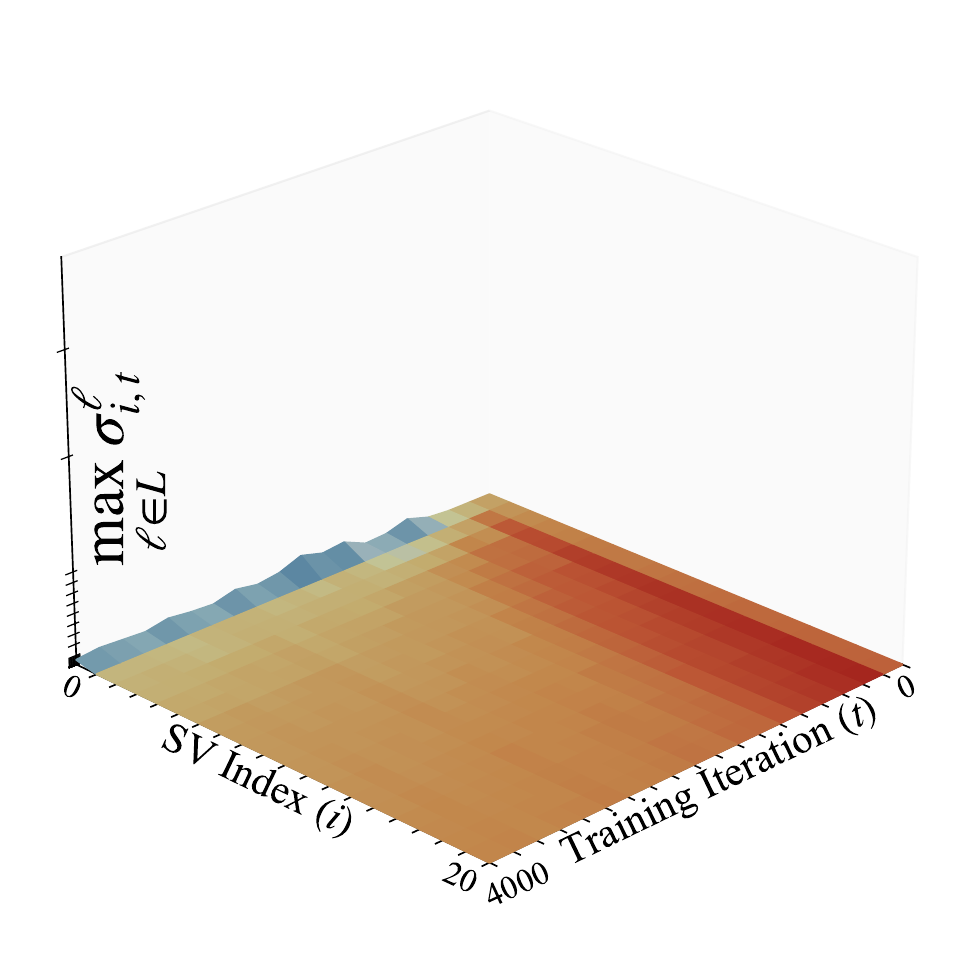}
        \caption{\scriptsize MLP-Up Proj.}
    \end{subfigure}
    \begin{subfigure}{0.24\textwidth}
        \centering
        \includegraphics[width=\linewidth, trim=25 25 20 40, clip]{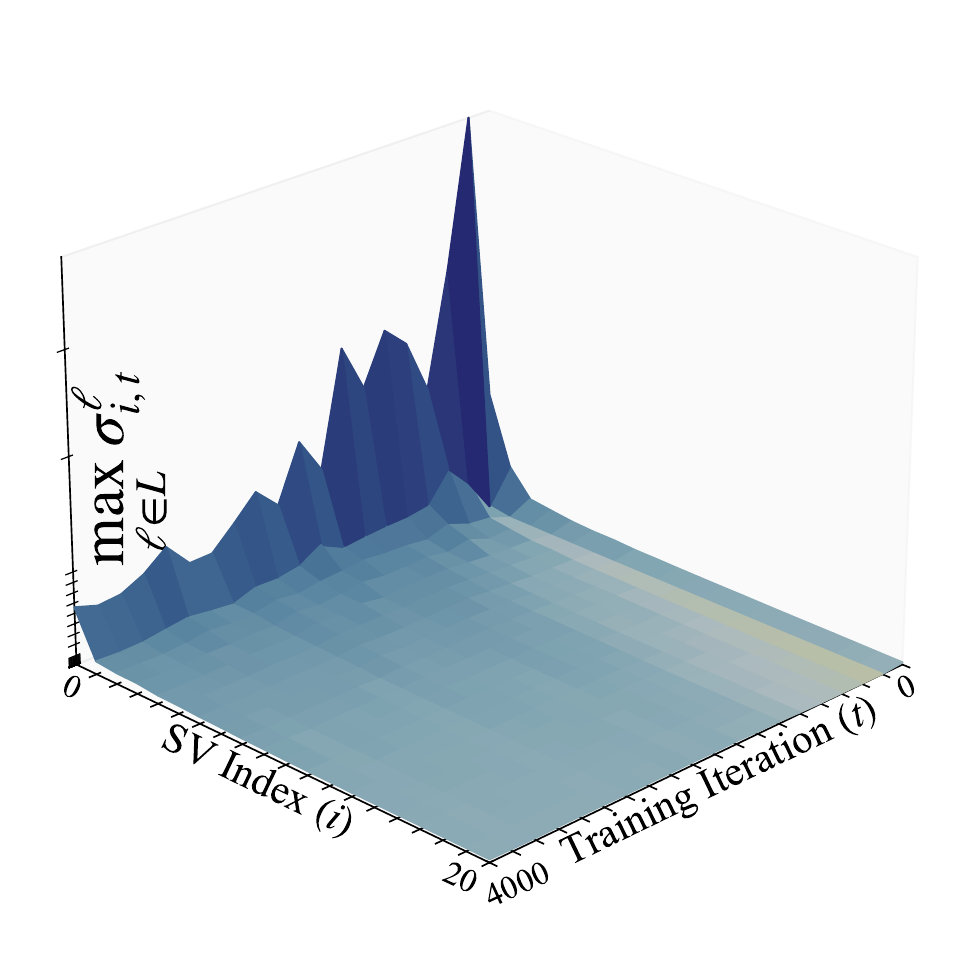}
        \caption{\scriptsize MLP-Down Proj.}
    \end{subfigure}
    \begin{subfigure}{0.24\textwidth}
        \centering
        \includegraphics[height=0.9\linewidth, width=0.5\linewidth, trim=0 0 0 0, clip]{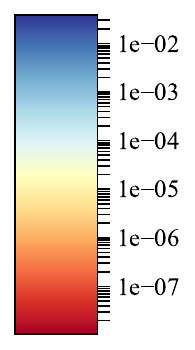}
        \caption{\scriptsize Legend}
    \end{subfigure}

    \caption{Evolution of the top singular values of the subspace estimation error derivative (${\partial F}/{\partial S_t}$) in the LLaMA-1B architecture. Each plot shows $\max \sigma_{i, t}^\ell$, the maximum \(i\)-th singular value across all layers $\ell$ within each layer type $L$, as training progresses. While MLP down-projection layers (g) exhibit the largest singular values, their magnitude remains small and decays rapidly. Other projection layers (a-d) show values close to zero throughout training. This suggests that the error spreads across all directions, and almost every subspace should perform similar.}
    \label{fig:singular_value_distribution}
\end{figure*}

%% file: Images/gradient_energy/tex.tex
\begin{figure*}[t!]
 % First Row

 \scriptsize
    \centering
    \begin{subfigure}{0.24\textwidth}
        \centering
        \includegraphics[width=\linewidth]{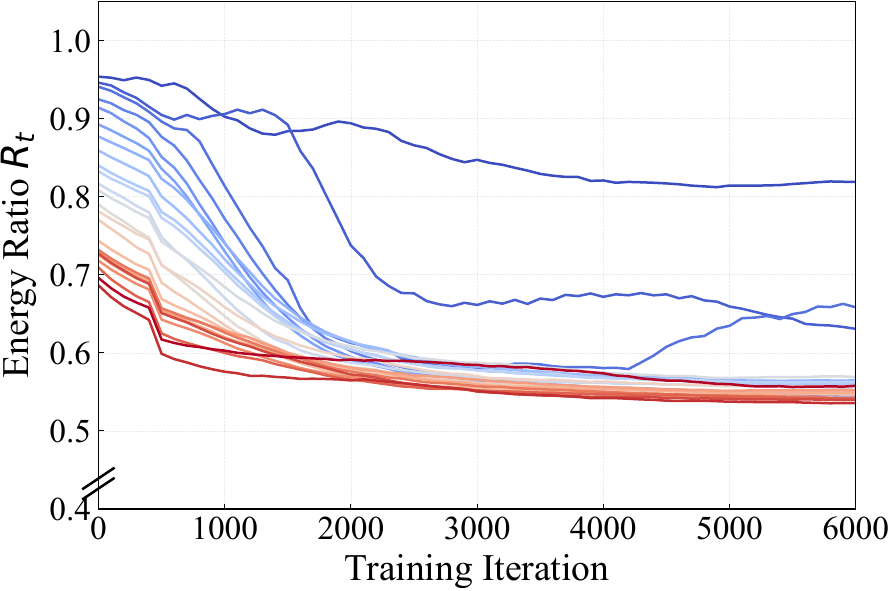}
        \caption{\scriptsize Attention-Output Proj.}
    \end{subfigure}
    \begin{subfigure}{0.24\textwidth}
        \centering
        \includegraphics[width=\linewidth]{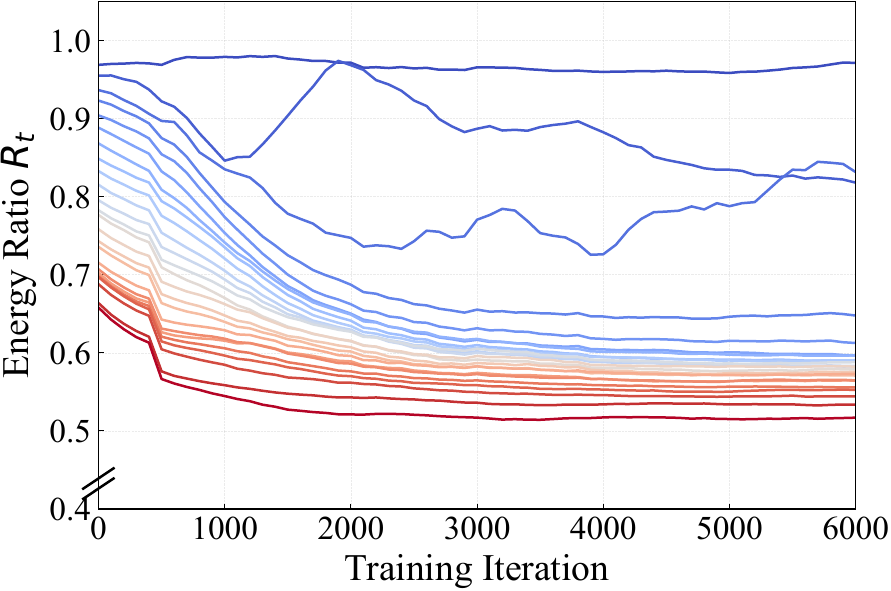}
        \caption{\scriptsize Attention-Value Proj.}
    \end{subfigure}
    \begin{subfigure}{0.24\textwidth}
        \centering
        \includegraphics[width=\linewidth]{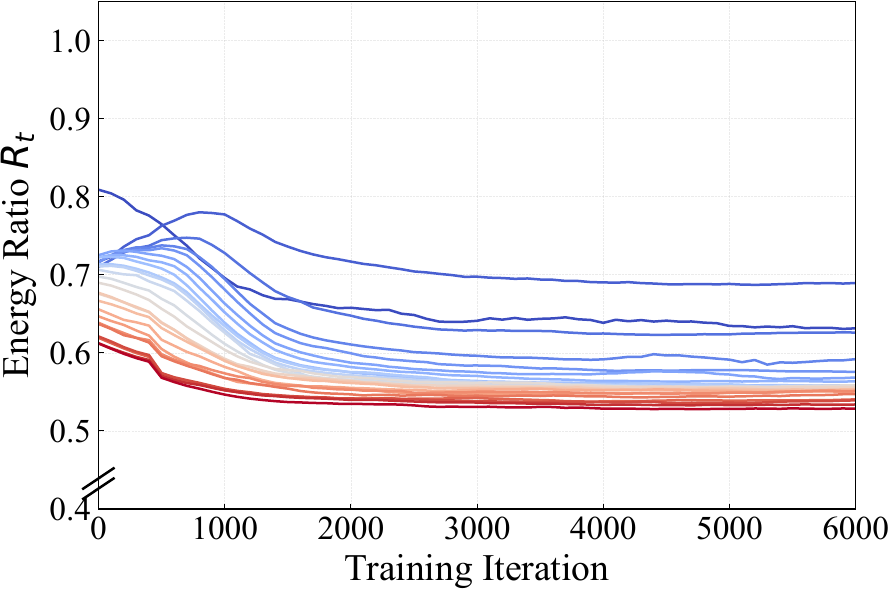}
        \caption{\scriptsize Attention-Query Proj.}
    \end{subfigure}
    \begin{subfigure}{0.24\textwidth}
        \centering
        \includegraphics[width=\linewidth]{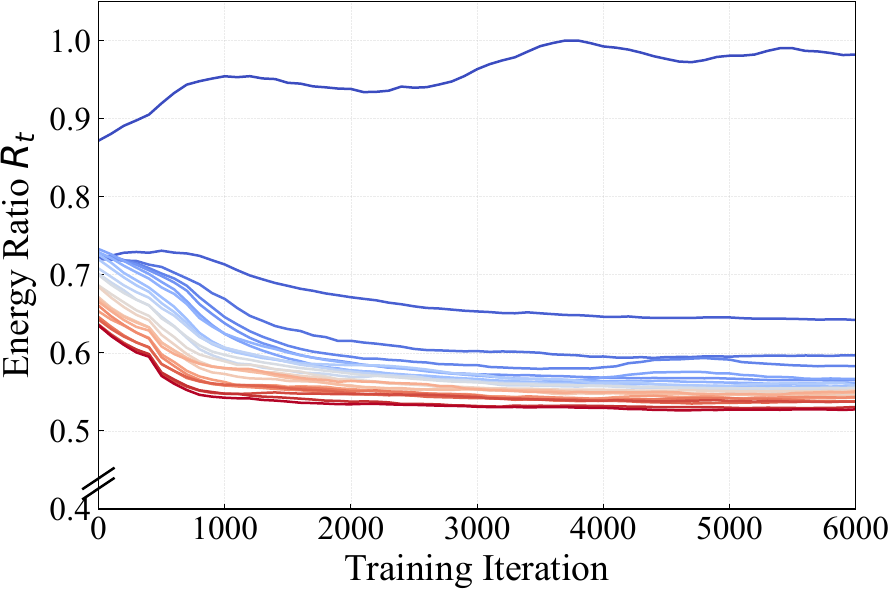}
        \caption{\scriptsize Attention-Key Proj.}
    \end{subfigure}

    % Second row
    \begin{subfigure}{0.24\textwidth}
        \centering
        \includegraphics[width=\linewidth]{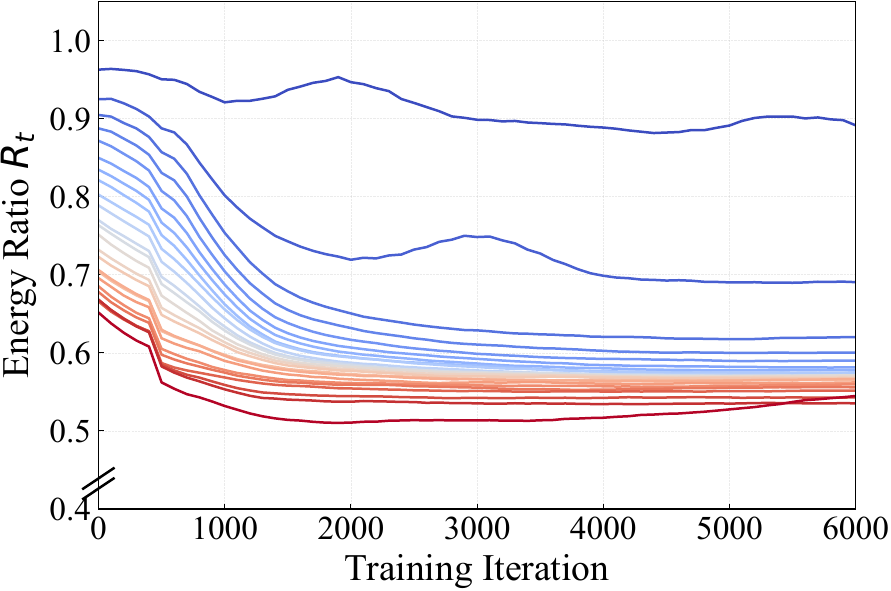}
        \caption{\scriptsize MLP-Gate Proj.}
    \end{subfigure}
    \begin{subfigure}{0.24\textwidth}
        \centering
        \includegraphics[width=\linewidth]{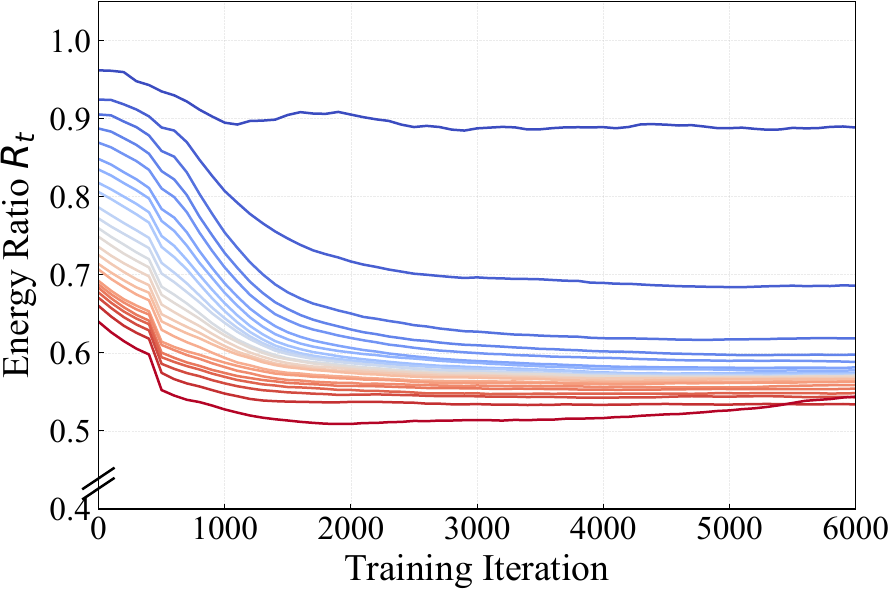}
        \caption{\scriptsize MLP-Up Proj.}
    \end{subfigure}
    \begin{subfigure}{0.24\textwidth}
        \centering
        \includegraphics[width=\linewidth]{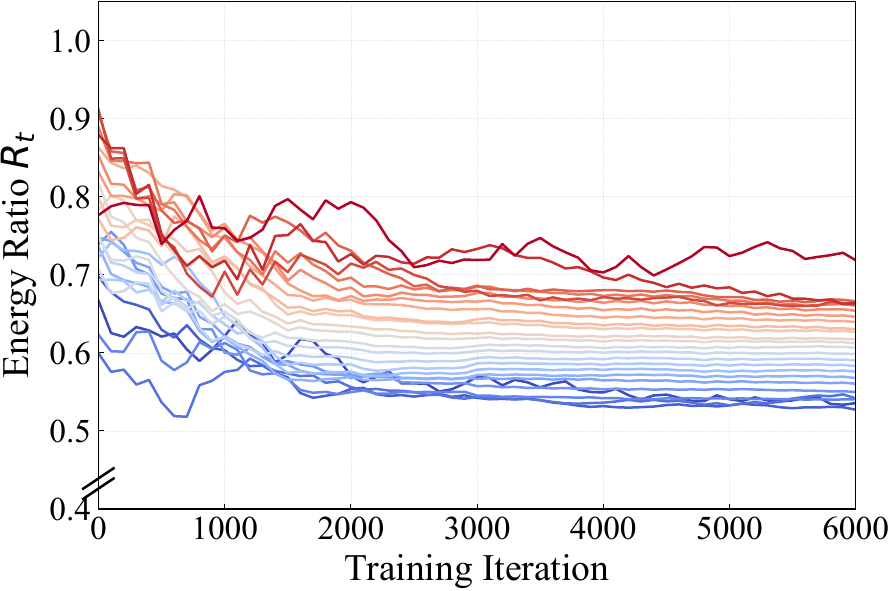}
        \caption{\scriptsize MLP-Down Proj.}
    \end{subfigure}
    \begin{subfigure}{0.24\textwidth}
        \centering
        \includegraphics[width=\linewidth]{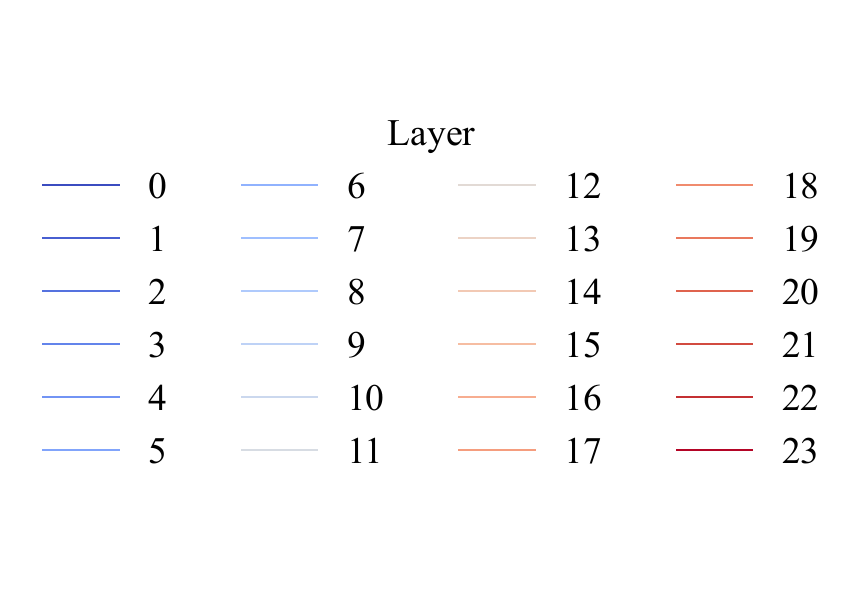}
        \caption{\scriptsize Legend}
    \end{subfigure}
    \caption{Each decoder layer stack includes seven layer types in the Llama-1B model. The plots show the fraction of gradient energy explained by a rank 512 approximation ($R_t = {\|\widetilde G_t}\|_F/{\|G_t\|_F}$). Despite a high lower bound, this fraction declines over training.}
    \label{fig:energy_fraction}
\end{figure*}

%% file: Images/ablation/tex.tex
% \begin{figure}[t]
\begin{wrapfigure}{r}{0.5\textwidth}
 % First Rown
    \centering
        \centering
        \includegraphics[width=\linewidth]{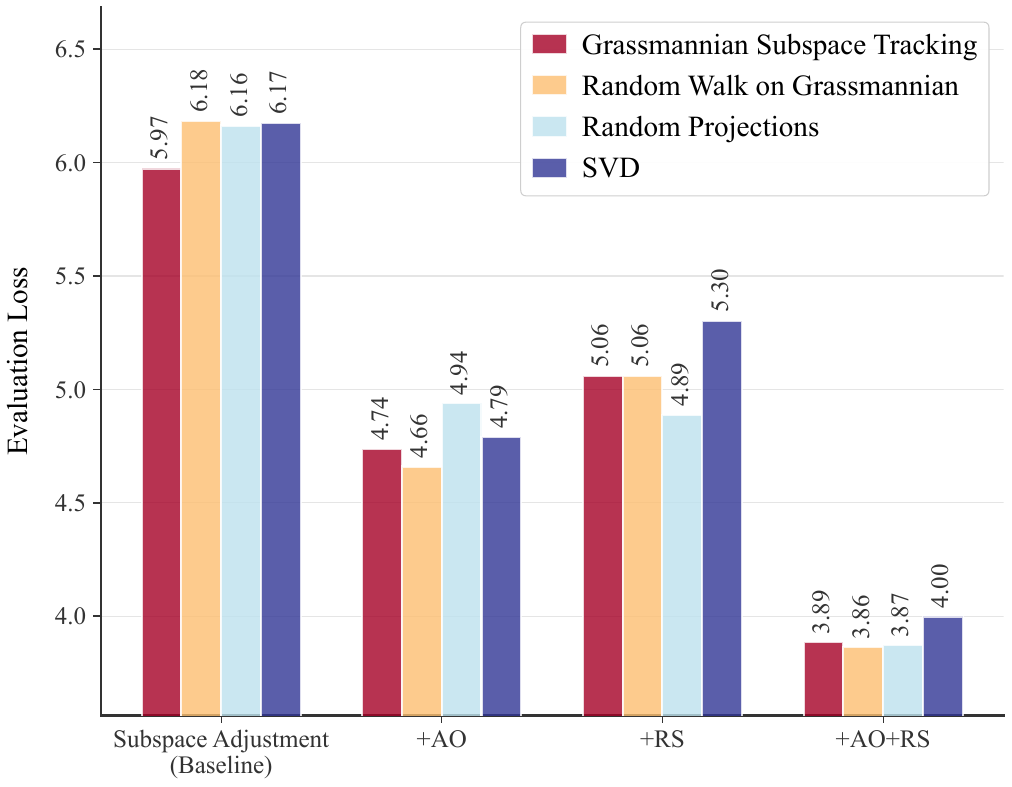}

    \caption{We ablate (i) the subspace update methods: Grassmannian subspace tracking, random walk on Grassmannian, random projections, and SVD; (ii) adaptive-optimizer (AO), and (iii) recovery scaling (RS), reporting evaluation loss.}
    \label{fig:ablation}
\vspace{-15pt}
% \end{figure}
\end{wrapfigure}

%% file: Sections/method.tex
\section{Random but Right}
\label{sec:method}
Building on our analysis of gradient subspace dynamics, we introduce two randomized methods, {\bf \methodname} and {\bf GrassJump}. These methods generate diverse subspaces through random exploration, preventing the optimizer from being trapped in flat regions of the manifold, and at the same time, they effectively incorporate solutions for the critical challenges of low-rank gradient methods.

{\bf Subspace Adjustment.} \methodname and GrassJump, differ in how they adjust the subspace. In GrassWalk, we first initialize the subspace from the SVD of the first gradient matrix. To update the subspace every $T$ iterations, we employ the exponential map on the Grassmannian manifold \citep{Bendokat_2024}, moving in a random direction from the current subspace. Concretely, we sample a random matrix $\mathbb X \in \mathbb R^{m \times r}$ to define the update direction in the tangent space. The update rule in \eqref{eq:update-rule} requires the SVD of $\mathbb X$ to move along the corresponding geodesic. Since we employ random directions, we approximate this decomposition using randomized SVD to reduce computational cost, denoting the result as $\small \widehat U_X \widehat\Sigma_X \widehat V_X^\top$. Here, $\tau$ denotes the update step size.
\begin{equation}
\small
\begin{aligned}
    S_{t+1}(\tau) = \;& S_t \widehat{V}_X \cos{(\widehat{\Sigma}_X\tau)} \widehat{V}_X^\top + \widehat{U}_X \sin{(\widehat{\Sigma}_X\tau)} \widehat{V}_X^\top  + S_t(I - \widehat{V}_X\widehat{V}_X^\top)
\end{aligned}
\label{eq:update-rule}
\end{equation}
For GrassJump, we adopt fully random projection matrices, effectively jumping from one point on the Grassmannian to another every $T$ iterations. At each update, we generate a random Gaussian matrix that can almost preserve orthogonality, especially when the dimensions are large.

% \textcolor{red}{Our analyses motivate the use of random projections for their two key benefits: Exploration and Generalization. The significant energy in the orthogonal space implies crucial information is often overlooked. Random projections act as a forcing exploration beyond dominant features. Also, in a flat curvature, random projections facilitate escaping local minima traps, thereby improving the generalization. Our methods, GrassJump and GrassWalk, are inspired by these benefits. GrassJump utilizes sudden subspace changes for aggressive exploration. In contrast, GrassWalk employs controlled, rank-1 random steps to escape spurious minima trap without sudden changes. Interestingly both methods can gain comparable performance; however, based on our ablations it seems distinct properties can offer different advantages: GrassWalk shows greater robustness to frequent updates Table \ref{tab:t-abl} as the adjustments are more controlled, whereas GrassJump performs better with lower ranks Table \ref{tab:r-abl} due to their independence to a core subspace. }

{\bf Informing the Optimizer of Subspace Updates.}
A key factor in adapting gradient low-rank methods is properly adapting the optimizer states when the subspace shifts. This adaptation becomes especially critical for leveraging the power of stateful optimizers, as demonstrated in Figure \ref{fig:ablation}. At each subspace update, we first transfer the previous first and second moments $M_{t-1}$ and $V_{t-1}$ into the new subspace, and then update them using the gradient at step $t$. For more details, please refer to Appendix \ref{app:ao-rs}.  
\begin{equation}
\small
    M_t \leftarrow \beta_1 (S_t^\top S_{t-1} M_{t-1}) + (1-\beta_1)\widetilde G_t
\label{eq:1st-moment-ao}
\end{equation}
\begin{equation}
\small
    V_t \leftarrow \beta_2\left[(1-\beta_2^{t-1})|(S_t^\top S_{t-1})^{2}\cdot (V_{t-1}-M_{t-1}^{2}) + (S_t^\top S_{t-1}\cdot M_{t-1})^{2}|\right] + (1-\beta_2)\widetilde G_t^{\,2}.
\label{eq:2nd-moment-ao}  
\end{equation}
{\bf Recovering Information Lost in Low-Rank Projections.}
We also recover the lost information $\Delta_t = G_t - S_t \widetilde G_t$ through a channel-wise rescaling mechanism. Specifically, we extract the scaling factor for each channel $j$ of the gradient, from the low-rank optimizer's output ($\widetilde{G}_t^O$), and the low-rank gradient ($\widetilde G_t$), as shown in \eqref{eq:phi}, and use that to scale the $j$-th channel of $\Delta_t$ for updating the weights. We detail the exact implementation of this recovery scaling in Appendix \ref{app:ao-rs}, and the pseudo code of \methodname and \secondname is presented in Algorithm \ref{alg:algo}.
\begin{equation}
\small
    \phi_t(\widetilde{G}_t)_i = \frac{\|\widetilde{G}^{i\,O}_{t}\|_2}{\|\widetilde{G}^i_{t}\|_2}
\label{eq:phi}
\end{equation}
The channel-wise treatment of gradients has been explored previously in the literature. Specifically, Apollo \citep{zhu2025apollosgdlikememoryadamwlevel} proposes reducing optimizer memory by maintaining a single scaling factor per channel rather than fine-grained, element-wise statistics. They theoretically prove that this approximation remains bounded when computed via random low-rank projections. Building on this channel-wise perspective, FiRA \citep{chen2025fira} and SubTrack++ \citep{rajabi2025subtrack} apply channel-wise scaling factors, derived directly from the low-rank optimizer's state, to rescale and recover the discarded orthogonal information after projection, however, they adopted structured subspace updates via SVD \citep{chen2025fira} and subspace tracking \citep{rajabi2025subtrack}. 

Here we demonstrate how the channel-wise scaling factors obtained from a low-rank optimizer serve as proportionally bounded estimators of their full-rank counterparts. This relationship formally justifies why we can effectively rescale and reintroduce the gradient information lost during low-rank truncation, even when utilizing random Gaussian projections. 
\input{Theories/recovery_scaling}
The proof of this theorem is provided at Appendix \ref{app:theory-rs}.

While Apollo \citep{zhu2025apollosgdlikememoryadamwlevel} demonstrated that element-wise scaling factors can be effectively replaced by channel-wise approximations, our approach here is different. We leverage the fine-grained optimizer within the low-rank subspace, applying these derived channel-wise scaling factors only to recover the orthogonal signal. As established by our theoretical bounds, this random projection-based recovery preserves the full-rank optimzer's behavior with high probability. Furthermore, this stateful recovery mechanism offers a distinct architectural advantage over methods like FRUGAL \citep{zmushko2025frugal}, which fall back on less accurate, stateless optimizers to handle residual signals. This theoretical superiority translates directly to the empirical performance gains demonstrated in Table \ref{tab:lama1b}.

%% file: Theories/recovery_scaling.tex
\begin{restatable}[\bf ‌Bounded Channel-Wise Scaling Factor]{theorem}{bounded}
\label{th:bounded}
The channel-wise ratio of the gradient scaling factor for the low-rank projected gradient using random projection $S$, sampled from a Gaussian distribution, to that of the original gradient is bounded as follows; for any channel \( j \), and with high probability:
\begin{equation}
\label{eq:main}
\small
    \frac{{1-\epsilon}}{(1+\epsilon)^2}\leq \sqrt{\frac{m}{r}}  \frac{\phi_t(\widetilde{G}_t)_j}{\phi_t(G_t)_j} \leq \frac{{1+\epsilon}}{(1-\epsilon)^2}.
\end{equation}
\end{restatable}

%% file: Sections/experiments.tex
\section{Experiments and Results}
We evaluated multiple baselines during pre-training LLaMA-1B and Qwen-1.5B architectures for 10K steps, using C4 dataset. All experiments are conducted on a single A6000 GPU, and the final evaluation loss is reported. All hyperparameters are reported in Appendix \ref{app:hyperparameters}.
Results presented in Tables \ref{tab:lama1b} show that \methodname and \secondname achieve state-of-the-art results. As illustrated in Figure \ref{fig:baselines}-a, the wall-clock time of our randomized methods matches that of the most computationally efficient methods, such as APOLLO \citep{zhu2025apollosgdlikememoryadamwlevel}, FRUGAL \citep{zmushko2025frugal} and SubTrack++ \citep{rajabi2025subtrack}, demonstrating superior convergence compared with other methods. 
\input{Tablels/llama1b_pretraining}

To assess generalization in larger models and for long-term trainings, we report the evaluation loss when pre-training a Llama-7B architecture up to 100K iterations in Table \ref{tab:lama7b}. In this experiments, \secondname outperforms \methodname and other baselines, highlighting the importance of thoroughly exploring the landscape, especially at long training regimes. Here, we exclude other baselines, as their performance was consistently weaker in Table \ref{tab:lama1b}.  Also, the training dynamics and loss during training are attached in Appendix \ref{app:long-run}, with hyperparameters reported in Appendix \ref{app:hyperparameters}.
\input{Tablels/llama7b_pretraining}

Table \ref{tab:t-abl} compares \methodname, GrassJump, and SubTrack++, while ablating subspace update interval.
Excessively frequent \secondname updates can degrade performance, suggesting sufficient iterations are required to stabilize local information, if updates are abrupt. Conversely, \methodname and SubTrack++ \citep{he2025subspace} show better performance under frequent updates due to their controlled, rank-1 adjustments, avoiding drastic changes. However, while GrassWalk with $T \in \{20, 50\}$ outperforms all the baselines in Table \ref{tab:lama1b}, SubTrack++ did not converge at T=20, highlighting the importance of random projections in flat optimization landscapes. Notably, omitting subspace updates entirely results in a final evaluation loss of 5.06, underscoring the necessity of subspace adaptation. Additional ablations investigating the sensitivity of GrassWalk and GrassJump are provided in Appendix \ref{app:ablations}.
\input{Tablels/t_abl}

%% file: Tablels/llama1b_pretraining.tex
% \begin{table}[h]
% \centering
% \caption{Comparison of low-rank methods on pretraining LLaMA-1B and {Qwen 1.5B} models. 
% We report evaluation loss (\(\downarrow\)), peak memory (GB), and wall-time (m). 
% Best results are in \textbf{bold}, and second best are \underline{underlined}.}
% \label{tab:lama1b}
% \begin{tabular}{lcccc}
% \toprule
% \textbf{Method} & \textbf{Architecture}& \textbf{Eval. Loss} (\(\downarrow\))  & \textbf{Peak Mem. (GB)} & \textbf{Wall Time (m)} \\
% \midrule
% GaLore \citep{zhao2024galorememoryefficientllmtraining}  
%     & Llama2-1B
%     & 6.17 & 31.1 & 522.2 \\
% APOLLO \citep{zhu2025apollosgdlikememoryadamwlevel}   
%     & Llama2-1B      
%     & 5.71 & 35.5 & 410.5 \\
% LDAdam \citep{robert2025ldadam}   
%     & Llama2-1B        
%     & 4.10 & 34.9 & 532.8 \\
% FRUGAL \citep{zmushko2025frugal}   
%     & Llama2-1B        
%     &  4.22   &  39.3   &  405.1 \\
% SubTrack++ \citep{rajabi2025subtrack}   
%     & Llama2-1B    
%     & 3.89 & 32.6 & 429.2 \\
% \midrule
% GrassWalk      
%     & Llama2-1B 
%     & \textbf{3.86} & 32.0 & 418.6 \\
% GrassJump   
%     & Llama2-1B 
%     & \underline{3.87} & 32.1 & 432.5 \\
%     \midrule
%     \midrule
% SubTrack++ \citep{rajabi2025subtrack}   
%     & {Qwen 1.5B}    
%     & {4.70} & {33.13} & {436.4} \\
% \midrule
% GrassWalk      
%     & {Qwen 1.5B}
%     & {\underline{4.68}} & {33.6} & {436.6} \\
% GrassJump   
%     & {Qwen 1.5B}
%     & {\textbf{4.66}} & {33.09} & {437.0} \\    
% \bottomrule
% \end{tabular}
% \vspace{-10pt}
% \end{table}

\begin{table*}[h!]
\centering
\caption{Comparing evaluation loss (\(\downarrow\)) on pretraining LLaMA-1B and {Qwen 1.5B} models, along with peak memory (GB), and wall-time (m). 
Best results are in \textbf{bold}, and second best are \underline{underlined}.}
\label{tab:lama1b}
\resizebox{\linewidth}{!}{
\begin{tabular}{c l c c c}
% \small
\toprule
\textbf{Arch.} & \textbf{Method} & \textbf{Eval. Loss}  
& \textbf{Peak Mem. (GB)} & \textbf{Wall Time (m)} \\
\midrule

% -------- LLaMA-1B block --------
\multirow{9}{*}{\rotatebox{90}{\small LlaMA-1B}} 
    & {AdamW [Full-Rank]} & {4.10} & {35.2} & {417.0} \\
    & GaLore \citep{zhao2024galorememoryefficientllmtraining}  
    & 6.17 & 31.1 & 522.2 \\
& APOLLO \citep{zhu2025apollosgdlikememoryadamwlevel}   
    & 5.71 & 35.5 & 410.5 \\
& LDAdam \citep{robert2025ldadam}   
    & 4.10 & 34.9 & 532.8 \\
& FRUGAL \citep{zmushko2025frugal}   
    & 4.22 & 39.3 & 405.1 \\
& SubTrack++ \citep{rajabi2025subtrack}   
    & 3.89 & 32.6 & 429.2 \\
& GrassWalk [Ours]      
    & \textbf{3.86} & 32.0 & 418.6 \\
% & GrassJump [Ours]   
%     & \underline{3.87} & 32.1 & 432.5 \\
& {GrassJump [Ours]}   
    & {\underline{3.87}} & {32.1} & {415.2} \\
\midrule\midrule

% -------- Qwen 1.5B block --------
\multirow{4}{*}{\rotatebox{90}{\small Qwen-1.5B}}
    &{AdamW [Full-Rank]} &{4.84} &{37.7} &{421.0} \\
    & {SubTrack++ \citep{rajabi2025subtrack}}    
    & {4.70} & {33.1} & {436.4} \\
& {GrassWalk [Ours]}       
    & {\underline{4.68}} & {33.6} & {436.6} \\
& {GrassJump [Ours]}   
    & {\textbf{4.67}} & {33.1} & {432.2} \\  

\bottomrule
\end{tabular}
}
% \vspace{-10pt}
\end{table*}

%% file: Tablels/llama7b_pretraining.tex
\begin{table*}[!h]
\centering
\small
\caption{Comparing evaluation loss ($\downarrow$) on pretraining LLaMA-7B, along with peak memory usage (GB), and wall-clock time (hours) after 10k and 100k of training iterations. Other baselines are omitted as their performance differs substantially from these three methods.
Best results are in \textbf{bold}.}
\label{tab:lama7b}
\resizebox{\linewidth}{!}{
\begin{tabular}{lccc}
\toprule
\textbf{Method} & \textbf{Eval. Loss - 10k} {\scriptsize(Wall Time)}& {\textbf{Eval. Loss - 100k} {\scriptsize(Wall Time)}} & \textbf{Peak Mem. (GB)}  \\
\midrule
% {GaLore \citep{zhao2024galorememoryefficientllmtraining}}   
%     & {6.23} {{\scriptsize(19.5 hours)}} & {3.36 {\scriptsize(139.1 hours)}} & {49.4}  \\
SubTrack++ \citep{rajabi2025subtrack}   
    & 4.94 {\scriptsize(9.57 hours)} & {3.36 {\scriptsize(93.2 hours)}} & 50.3  \\
GrassWalk [Ours]   
    & 4.94 {\scriptsize(9.55 hours)} & {3.37 {\scriptsize(93.2 hours)}} & 50.3 \\
GrassJump [Ours]
    & \textbf{4.86} {\scriptsize(9.24 hours)} & {\textbf{3.34} {\scriptsize(91.6 hours)}} & 48.7  \\
\bottomrule
\end{tabular}
}
\vspace{-10pt}
% \end{wraptable}
\end{table*}

%% file: Tablels/t_abl.tex
\begin{table}[!h]
\centering
\caption{Final evaluation loss (\(\downarrow\)) after pre-training a Llama-1B architecture for 10k iterations with $r = 512$ and with different subspace update intervals.}
\label{tab:t-abl}
\resizebox{\linewidth}{!}{
\begin{tabular}{lccccccc}
\toprule
\textbf{Method} & \textbf{T = 20} & \textbf{T = 50} & \textbf{T = 100} &\textbf{T = 200} &\textbf{T = 500} &\textbf{T = 1K} &\textbf{T = 2K} \\
\midrule
SubTrack++ \citep{rajabi2025subtrack} &  NC & 3.86 & 3.89 & 3.99 & 4.25 & 4.45 & 4.68 \\
GrassWalk [Ours]   
    &  3.72 & 3.79 & 3.86 & 3.98 & 4.25 & 4.46 & 4.68 \\
GrassJump [Ours]
    & 4.33 & 4.34 & 3.87 & 3.96 & 4.15 & 4.75 & 4.89 \\
\bottomrule
\end{tabular}
}
\vspace{-10pt}
\end{table}

%% file: Sections/related_works.tex
\section{Related Works}
LoRA \citep{hu2021lora} reduces memory via low-rank adapters, with extensions such as QLoRA \citep{dettmers2024qlora} and Deep LoRA \citep{yaras2024compressible} improving efficiency and robustness. Further variants enhance adaptation \citep{lialin2023relorahighranktraininglowrank, renduchintala-etal-2024-tied, xia2024chainloraefficientfinetuning, pan2024lisalayerwiseimportancesampling}, while other approaches boost memory efficiency by compressing activations \citep{miles2024veloramemoryefficienttraining} or reformulating optimization via block coordinate descent \citep{luo2024badammemoryefficientparameter}. FLora \citep{hao2024floralowrankadapterssecretly} provides a complementary perspective by showing that LoRA can be interpreted as a random projection gradient compressor. Another line of work exploits the structure of high-dimensional data by projecting it into evolving subspaces. Incremental and Grassmannian-based methods have been proposed for subspace tracking under partial observations \citep{balzano2011onlineidentificationtrackingsubspaces}, noise \citep{zhang2016globalconvergencegrassmanniangradient, kasai2017fastonlinelowranktensor}, and geodesic evolution \citep{blocker2023dynamicsubspaceestimationgrassmannian}, offering a principled foundation for projections in LLM training.

As optimizers like Adam \citep{kingma2017adammethodstochasticoptimization} account for a significant portion of memory, there are many methods \citep{modoranu2024microadamaccurateadaptiveoptimization}, \citep{zhang2024adamminiusefewerlearning} that aim to reduce optimizer states. MicroAdam \citep{modoranu2024microadamaccurateadaptiveoptimization} compresses gradients with feedback correction, while Adam-mini \citep{zhang2024adamminiusefewerlearning} partitions models into blocks with shared learning rates. 
\citep{gurari2018gradientdescenthappenstiny}, \citep{schneider2024identifyingpolicygradientsubspaces} show that a substantial portion of gradients lies within a largely consistent subspace. GaLore \citep{zhao2024galorememoryefficientllmtraining} first leverage this fact to reduce optimizer's memory by projecting gradients onto a low-rank subspace, yielding large memory savings. \citet{jaiswal2024galorewelorelowrankweights} fine-tune only layers with low-dimensional gradient subspaces, while Grass \citep{muhamed2024grasscomputeefficientlowmemory} saves memory via sparse gradient projections. \citet{ramesh2024blockllmmemoryefficientadaptationllms} achieve efficiency by dynamically updating only a subset of parameters. GoLore \citep{he2025subspace} addresses GaLore’s convergence issue and by injecting random projections in later iterations, ensures convergence. Fira \citep{chen2025fira} uses norm-based scaling to transfer the adaptive behavior of a low-rank optimizer to full-rank updates. 
APOLLO \citep{zhu2025apollosgdlikememoryadamwlevel} approximates channel-wise scaling using  random subspaces, effectively coarsening learning-rate adaptation with SGD-like memory. RSO \citep{chen2025a} decomposes training into sequences of randomized lower-dimensional subproblems. Adapprox \citep{Adapprox} targets Adam’s second moment with randomized low-rank approximations, adaptive rank, and similarity guidance. GreedyLore \citep{greedylore} explores greedy gradient compression with error-feedback and semi-lazy subspace updates. Also, FRUGAL \citep{zmushko2025frugal} leverages gradient splitting: it applies stateful updates in a low-dimensional space and state-free methods along remaining directions, via columnwise random projections.

%% file: Sections/conclusion.tex
\section{Limitations and Future Work}
While providing strong evidence for the efficacy of randomized optimization, several limitations present opportunities for future research. First, although our analysis suggest that the optimization is not sensitive to the choice of subspace, the phenomenon requires further exploration across more diverse settings, architectures, and tasks, and a more rigorous formal characterization of flatness will be critical. Second, due to the immense computational resources, our experiments were constrained to models of up to 7 billion parameters. Finally, while the theoretical guarantees provided in this paper align perfectly with standard assumptions in the literature, our mathematical bounds serve as a conservative theoretical proxy rather than a precise reflection of practical optimization trajectories.

%% file: Sections/appendix.tex
\section{Adaptive Optimizer and Recovery Scaling}
\label{app:ao-rs}
\textbf{Adaptive Optimizer. }In Adam, the momentum update rules compute weighted averages of the first- and second-order gradient moments using the parameters $\beta_1$ and $\beta_2$, as shown in the following equations.
\begin{equation}
\label{eq:M-adam}
\small
    M_t \gets \beta_1 \cdot M_{t-1} + (1 - \beta_1) \cdot \widetilde{G}_t 
\end{equation}
\begin{equation}
\label{eq:V-adam}
\small
    \mathcal{V}_t \gets \beta_2 \cdot \mathcal{V}_{t-1} + (1 - \beta_2) \cdot \widetilde{G}_t^2
\end{equation}
When the subspace is updated, we rotate Adam's moments onto the new basis so that the optimizer remains aligned with the updated subspace. Orthogonal projection works well for the first moment but not for the second, since Adam involves nonlinear operations. To handle this, we treat Adam’s states as statistical estimates of the first and second moments of each gradient coordinate, and thus using \eqref{eq:1st-moment-ao} and \eqref{eq:2nd-moment-ao} for our adaptive optimizer (AO). A similar perspective has also been adopted in prior state-of-the-art methods \citep{robert2025ldadam, rajabi2025subtrack}.
\begin{equation*}
\small
    M_t \leftarrow \beta_1 (S_t^\top S_{t-1} M_{t-1}) + (1-\beta_1)\widetilde G_t
\end{equation*}
\begin{equation*}
\small
    V_t \leftarrow \beta_2\left[(1-\beta_2^{t-1})|(S_t^\top S_{t-1})^{2}\cdot (V_{t-1}-M_{t-1}^{2}) + (S_t^\top S_{t-1}\cdot M_{t-1})^{2}|\right] + (1-\beta_2)\widetilde G_t^{\,2}.
\end{equation*}

\textbf{Recovery Scaling. }Based on the observation that the scale ratio between dominant and bulk subspaces is consistent \citep{zhu2025apollosgdlikememoryadamwlevel, chen2025fira}, we reintroduce this signal by columnwise rescaling of $\Delta_t$ according to the ratio between the optimizer’s output $\widetilde G_t^O$ and the raw low-rank gradient $\widetilde G_t$, as shown in \eqref{eq:rs}. This enables the use of stateful optimizer dynamics without storing the full optimizer states. With a growth-rate limiter $\zeta$, we prevent the scaling from diverging. Specifically, if $\|\Lambda_t\| / \|\Lambda_{t-1}\| > \zeta$, we rescale as per \eqref{eq:limiter}. Several works \citep{rajabi2025subtrack, zmushko2025frugal, chen2025fira, zhu2025apollosgdlikememoryadamwlevel, robert2025ldadam} have employed various recovery scaling (RS) techniques.
\begin{equation}
\label{eq:rs}
\small
    \phi_t(\widetilde G_t)_i = \frac{\|\widetilde G^{i\,O}_{t}\|}{\|\widetilde G^i_{t}\|},\qquad
    \Lambda_t = \phi_t(\widetilde G_t)\,\Delta_t,
\end{equation}
\begin{equation}
\label{eq:limiter}
\Lambda_t \leftarrow \Lambda_t \cdot \frac{\zeta \|\Lambda_{t-1}\|}{\|\Lambda_t\|}.
\end{equation}
The pseudo-code of \methodname \secondname are provided in Algorithm \ref{alg:algo}.
\input{Algorithm/algorithm}
\input{Images/baselines/tex}
\section{Geometry Preservation Proof}
\label{app:geometry-proof}
\input{Theories/geometry_preservation_proof}

\section{Bounded Channel-Wise Scaling Factor Proof}
\label{app:theory-rs}
\input{Theories/recovery_scaling_proof}

\section{ Core Subspace Energy vs. Rank}
\label{app:energy_frac}
In Figure \ref{fig:energy_fraction}, we illustrate the fraction of gradient energy captured by the core subspace, as defined in \eqref{eq:fraction}. These plots were generated during pre-training of a Llama-1B architecture using a subspace rank of $r = 512$. To further examine this phenomenon across different subspace capacities, we additionally include results for ranks 256 (Figure \ref{fig:energy_fraction-256}) and 1024 (Figure \ref{fig:energy_fraction-1024}).

As anticipated, increasing the rank also increases the energy fraction. Crucially, even with rank 1024, a considerably high rank for a 1B model, the fractions stabilize between 70\% and 80\% after initial iterations (with a significant 20\% to 30\% scattered on the orthogonal space), consistently demonstrating the decay pattern across nearly all layer types, except for MLP-down projection. For rank 256, the fractions are significantly lower, settling around 40\% to 50\%. This extended experimentation strongly validates the conclusion drawn from Figure \ref{fig:energy_fraction}.

\input{Images/gradient_energy_256/tex}
\input{Images/gradient_energy_1024/tex}

\section{ Additional Ablations}
\label{app:ablations}
In this section we report the result of ablations against different hyperparameters.
\subsection{Subspace Rank}
The rank of the core subspace, when pre-training the Llama-1B architecture with a subspace update interval of 200, impacts the evaluation loss as shown in the Table \ref{tab:r-abl}. While a higher rank generally yields better loss, the marginal benefit is limited. Notably, in the GrassJump, the reduced dependency on a specific subspace allows for effective performance even with significantly lower ranks.
\input{Tablels/r_abl}

% \subsection{ Subspace Update Interval}
% Table \ref{tab:t-abl} summarizes the impact of update frequency on pre-training of aLlama-1B architecture with rank 512. Notably, no subspace update results in a final evaluation loss of 5.06 (Figure 3), underscoring the necessity of subspace adaptation. Crucially, excessively frequent GrassJump updates degrade performance, suggesting that drastic subspace changes disrupt the optimizer's effectiveness and necessitate sufficient iteration for convergence.

\subsection{ Norm-Growth Limiter}
\input{Images/norm_growth_limiter/tex}
In Figure \ref{fig:zeta}, we examine the effect of different norm-growth limiter values on the training dynamics of a Llama-1B model trained with rank 512 using the GrassWalk and GrassJump methods. As shown, this parameter acts similarly to a gradient-clipping coefficient, regulating training spikes. Furthermore, Table \ref{tab:z-abl} demonstrates that mitigating these spikes, while avoiding excessive suppression of the recovered signal (e.g., ($\zeta = 0.1$)) has a substantial impact on the final evaluation loss.
\input{Tablels/z_abl}

\subsection{ The Step-Size of the Grassmannian Updates}
GrassJump employs purely random subspace selection for maximal exploration. Conversely, GrassWalk incorporates an $\eta$ parameter, representing the step-size for updates on the Grassmannian manifold that can affect its performance. Ablation studies indicate that the step-size is inconsequential to the final pre-training evaluation loss of GrassWalk on pre-training a Llama-1B model, as demonstrated in Table \ref{tab:e-abl}. Notably, the step-size's appearance in the sine and cosine terms of \eqref{eq:update-rule} inherently leads to oscillatory behavior after scaling. 
\input{Tablels/e_abl}

\section{ Long-Run Training of 7B Model}
\label{app:long-run}
We pre-trained the Llama-7B architecture using GrassWalk, GrassJump, and SubTrack++ \citep{rajabi2025subtrack} for 100K steps. The results are presented in Table \ref{tab:lama7b}, and Figure \ref{fig:long-run} shows GrassJump consistently outperforms both in training dynamics and evaluation loss. Notably, GrassWalk achieves comparable performance without requiring gradient tracking.
\input{Images/long_run_pretraining/tex}

\section{ Pre-Training Hyperparameters}
\label{app:hyperparameters}
The hyperparameters of the experiments are reported in Table \ref{tab:pt_hyperparameters}. All experiments are conducted on A6000 GPUs. 
\input{Tablels/hyperparameters}

\section{ Flat Curvature}
\label{app:flat-curve}
Figure \ref{fig:singular_value_distribution} shows the distribution of the singular values of the tangent vectors of the subspace-estimation error across layers and training iterations. Figure \ref{fig:tangent-fro} plots the Frobenius norm of these singular values for each layer over the course of training. As illustrated, nearly all layers exhibit extremely small values, indicating that the subspace-estimation error has very low sensitivity in almost every direction.
\input{Images/tangent_fro/tex}

%% file: Algorithm/algorithm.tex
\begin{algorithm*}[h]
\caption{\\
\textbf{\textcolor{NavyBlue}{GrassWalk}} and \textbf{\textcolor{BrickRed}{GrassJump}}}
\label{alg:algo}

\begin{algorithmic}
\footnotesize
\REQUIRE \(W_t\), \(G_t \in \mathbb{R}^{m \times n}\) with \(m \leq n\) (w.l.o.g.), learning rate \(\alpha\), decay rates \(\beta_1\) and \(\beta_2\), subspace update interval \(T\), recovery scaling limiter factor \(\zeta\).
\STATE 
    {\textcolor{NavyBlue}{GrassWalk:}} \(S_0 \gets U[:, :r]\) , where \(U, S, V \gets \text{SVD}(G_0)\) \hfill \COMMENT{initialing the subspace}\\
    {\textcolor{BrickRed}{GrassJump:}} \(S_0 \gets\) random gaussian matrix 
\WHILE{Convergence}
    \IF{\(step\) mod \(T == 0\)} 
    \STATE
    {\textcolor{NavyBlue}{GrassWalk:}} \(S_{t}(\tau) = S_{t-1} \widehat{V}_X \cos{(\widehat{\Sigma}_X\tau)} \widehat{V}_X^\top + \widehat{U}_X \sin{(\widehat{\Sigma}_X\tau)} \widehat{V}_X^\top + S_{t-1}(I - \widehat{V}_X\widehat{V}_X^\top)\), \hfill \COMMENT{updating the subspace} \\ 
    where \(\widehat{U}_X, \widehat{\Sigma}_X, \widehat{V}_X^\top \gets \) SVD(random gaussian matrix)   \\
        {\textcolor{BrickRed}{GrassJump:}} \(S_t \gets \) random gaussian matrix \\
        \(\widetilde{G}_t = S_t^\top G_t\) \hfill \COMMENT{calculate low-rank gradient}
    \STATE
    \(M_t \leftarrow \beta_1 (S_t^\top S_{t-1} M_{t-1}) + (1-\beta_1)\widetilde G_t\) \hfill \COMMENT{toggle adaptive optimizer} \\
    \(V_t \leftarrow \beta_2\left[(1-\beta_2^{t-1})|(S_t^\top S_{t-1})^{2}\cdot (V_{t-1}-M_{t-1}^{2}) + (S_t^\top S_{t-1}\cdot M_{t-1})^{2}|\right] + (1-\beta_2)\widetilde G_t^{\,2}.\) \\
    \ELSE
    \STATE \(S_t = S_{t-1}\)
    \STATE
    \(M_t \gets \beta_1 \cdot M_{t-1} + (1 - \beta_1) \cdot \widetilde{G}_t \) \hfill \COMMENT{regular optimizer}\\
    \(\mathcal{V}_t \gets \beta_2 \cdot \mathcal{V}_{t-1} + (1 - \beta_2) \cdot \widetilde{G}_t^2\)
    \ENDIF
    \STATE 
    \(\widetilde{G}_t^O \gets \) optimizer's output, \(\widehat{G}_t = S_t \widetilde{G}_t^O\)
    \STATE 
    \(\phi_t(\widetilde G_t)_i = \frac{\|\widetilde G^{i\,O}_{t}\|}{\|\widetilde G^i_{t}\|},\qquad
    \Lambda_t = \phi_t(G_t)\,\Delta_t, \) \hfill \COMMENT{recovery scaling}\\
    \(\Lambda_t \leftarrow \Lambda_t \cdot \frac{\zeta \|\Lambda_{t-1}\|}{\|\Lambda_t\|}\)
        \STATE \(W_t \gets W_{t-1} - \alpha \cdot \widehat{G}_t\) \(- \alpha \cdot \Lambda_t\) \hfill \COMMENT{weight update rule}
\ENDWHILE
\end{algorithmic}
\end{algorithm*}

%% file: Images/baselines/tex.tex
\begin{figure*}[t]
 % First Rown
 \small
    \centering
    \begin{subfigure}{0.49\textwidth}
    \label{fig:1b-}
        \centering
        \includegraphics[width=\linewidth]{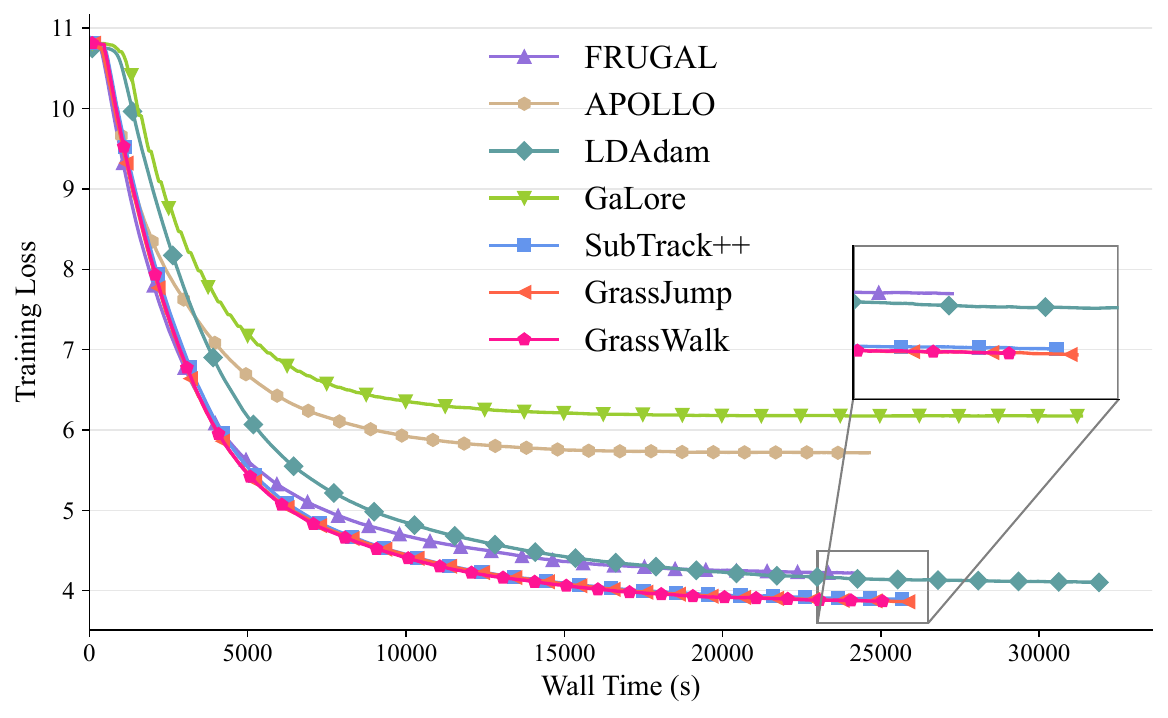}
        \caption{\scriptsize Pre-training Llama-1B}
    \end{subfigure}
    \begin{subfigure}{0.49\textwidth}
    \label{fig:7b-}
        \centering
        \includegraphics[width=\linewidth]{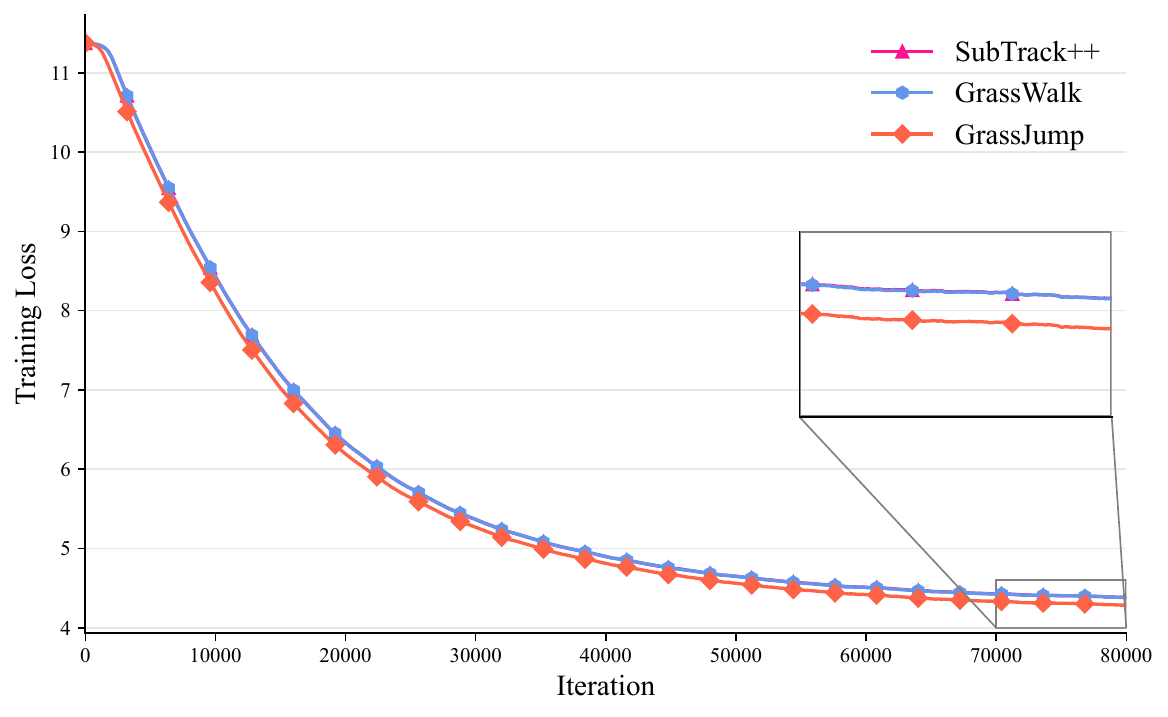}
        \caption{\scriptsize Pre-training Llama-7B}
    \end{subfigure}
    \caption{Comparison of different methods on LLaMA pretraining. (a) Wall-clock training curves for LLaMA-1B across all baselines. (b) Pretraining results for LLaMA-7B across selected methods, excluding weaker baselines due to their large performance gap.}
    \label{fig:baselines}
\end{figure*}

%% file: Theories/geometry_preservation_proof.tex
\normpreservation*
For providing the proof, we will first define the following two terms:
\begin{definition}[\bf Unit Sphere]
For $m \ge 1$, the unit sphere in $\mathbb{R}^m$ is defined by
\begin{equation*}
\small
\mathbb{S}^{m-1}
:=
\left\{x \in \mathbb{R}^m : \|x\|_2 = 1\right\}.
\end{equation*}
\end{definition}

\begin{definition}[\bf Lipschitz Norm]
Let $f:\mathbb{S}^{m-1}\to \mathbb{R}$ be a function. The Lipschitz norm of $f$ is defined by
\begin{equation*}
\small
\|f\|_{\mathrm{Lip}}
:=
\sup_{\substack{x,y\in \mathbb{S}^{m-1}\\ x\neq y}}
\frac{|f(x)-f(y)|}{\|x-y\|_2}.
\end{equation*}
Equivalently, $f$ is Lipschitz with constant $L$ if
\begin{equation*}
\small
|f(x)-f(y)| \le L\|x-y\|_2
\qquad
\text{for all } x,y\in \mathbb{S}^{m-1},
\end{equation*}
and $\|f\|_{\mathrm{Lip}}$ is the smallest such constant $L$.
\end{definition}

{\bf Proof.} Assuming $\|x\|_2 = 1$ without loss of generality, we examine an equivalent setup. By treating the projection $S$ as fixed and the vector $x$ as random ($x \in Unif(\mathbb{S}^{m-1})$), the underlying distribution of $\|S^\top x\|_2$ is perfectly preserved (using rotation invarice).

Again, without loss of generality, we can assume that $S$ is the
coordinate projection onto the first $r$ coordinates in $\mathbb{R}^m$. Thus
\begin{equation*}
\small
\mathbb{E}\,\|S^\top x\|_2^2 = \mathbb{E}\sum_{i=1}^{r} x_i^2 =
\sum_{i=1}^{r}\mathbb{E}x_i^2 = r\mathbb{E}x_1^2
\end{equation*}
since the coordinates $x_i$ of the random vector $x$ are identically distributed. To compute $\mathbb{E}x_1^2$, note that we assumed:
\begin{equation*}
\small
1 = \|x\|_2^2 = \sum_{i=1}^{m} x_i^2.
\end{equation*}
Taking expectations of both sides, we obtain
\begin{equation*}
\small
1 = \sum_{i=1}^{m}\mathbb{E}x_i^2 = m\mathbb{E}x_1^2,
\end{equation*}
which yields
\begin{equation*}
\small
\mathbb{E}x_1^2 = \frac{1}{m}.
\end{equation*}
And thus:
\begin{equation*}
\small
\mathbb{E}\,\|S^\top x\|_2^2 = \frac{r}{m}.
\end{equation*}
Now, for proving the probability, will use the following Lemma:

\begin{lemma}[\bf Concentration for the Unit Sphere]
Let $X \sim \operatorname{Unif}(S^{m-1})$, and let $f:\mathbb{S}^{m-1}\to \mathbb{R}$ be a Lipschitz function. Then
\begin{equation*}
\small
\|f(X)-\mathbb{E}f(X)\|_{\psi_2}
\le
\frac{C\|f\|_{\mathrm{Lip}}}{\sqrt{m}}.
\end{equation*}
where $\|\cdot\|_{\psi_2}$ is the sub-gaussian norm. Equivalently, for every $t \ge 0$,
\begin{equation*}
\small
\mathbb{P}\left\{\left|f(X)-\mathbb{E}f(X)\right|\ge t\right\}
\le
2\exp\left(
-\frac{c m t^2}{\|f\|_{\mathrm{Lip}}^2}
\right).
\end{equation*}
\end{lemma}
For checking the proof of this Lemma, please refer to \cite{Vershynin_2018}.

Using this lemma, we now define $f(x) := \|S^\top x\|_2$, which is a Lipschitz function on $\mathbb{S}^{m-1}$, and $\|f\|_{\mathrm{Lip}} = 1$ (check \cite{Vershynin_2018}).
Then:
\begin{equation*}
\small
\mathbb{P}\left\{
\left|\|S^\top x\|_2 - \sqrt{\frac{r}{m}}\|S^\top x\|_2\right| \ge t
\right\}
\le
2\exp(-cmt^2).
\end{equation*}
Choosing
$t := \varepsilon\sqrt{r/m}$ completes the proof. $\blacksquare$

Now, we get to prove the next theorem:
\distancepreservation*

{\bf Proof.}
Consider the following difference set
\begin{equation*}
\small
X-X := \{x-y : x,y\in X\}.
\end{equation*}
We need to show that the following inequality holds, with the required probability in the theorem:
\begin{equation*}
\small
(1-\varepsilon)\|z\|_2 \le \|S'^\top z\|_2 \le (1+\varepsilon)\|z\|_2
\end{equation*}
holds for all $z\in X-X$. Since $S'=\sqrt{\frac{m}{r}}\,S$, this inequality is equivalent to
\begin{equation*}
\small
(1-\varepsilon)\sqrt{\frac{r}{m}}\,\|z\|_2
\le
\|S^\top z\|_2
\le
(1+\varepsilon)\sqrt{\frac{r}{m}}\,\|z\|_2.
\end{equation*}
In Theorem \ref{th:norm-preservation}, we showed that for any fixed $z$ this holds with probability at least
$1 - 2\exp(-c\varepsilon^2 r)$. Now, we need to take a union bound over
$z\in X-X$. Given that for all $z\in X-X$, we have the stated probability, we can derive:
\begin{equation*}
\small
1 - |X-X|\,2\exp(-c\varepsilon^2 r)
\ge
1 - 2n^2 \exp(-c\varepsilon^2 r).
\end{equation*}
We replace $n^2$ by its exponential form:
\begin{equation*}
\small
1 - |X-X|\,2\exp(-c\varepsilon^2 r)
\ge
1 - 2\exp(2 \log n) \exp(-c\varepsilon^2 r) = 1 - 2\exp(2 \log n -c\varepsilon^2r).
\end{equation*}
If
\begin{equation*}
\small
r \ge \frac{C}{\varepsilon^2}\log n,
\end{equation*}
then,
\begin{equation*}
    \small
    \log n \le \frac{\varepsilon^2}{C}r
\end{equation*}
and if we set $C$ large enough, we can lower the probability, specifically, if $C = 4/c$, then the probability of failure will be at most $2\exp(-c\varepsilon^2r/2)$, and the theorem is proved as claimed. $\blacksquare$ For more details and background, one can check \cite{Vershynin_2018}.

%% file: Theories/recovery_scaling_proof.tex
Before starting proof of this Theorem \ref{th:bounded}, we will start by stating two lemmas, and providing their proof, based on the theoretical analysis provided on Apollo \citep{zhu2025apollosgdlikememoryadamwlevel}, and concepts of randomized linear algebra \citep{Vershynin_2018, Woodruff_2014}:

\begin{restatable}[\bf Channel-Wise First Momentum Ratio Norm Is Bounded]{lemma}{firstmomentum}
\label{th:first-momentum}
Let $G_t \in \mathbb{R}^{m \times n}$ be the full-rank gradient, and $S$ be a matrix of shape $\mathbb{R}^{m \times r}$, sampled from a Gaussian distribution in the variance of $1/r$. 
With the projected gradient $\widetilde{G}_t = S^\top G_t$,
we have the projected gradient with a bounded channel-wise first order moment,
for any channel \( j \),
and with probability at least $1 - 2\exp\left(-cr\epsilon^2\right)$:
\begin{equation}
\small
(1-\epsilon)\|M_t^j\|_2 \leq \|\widetilde{M}^j_t\|_2 \leq (1+\epsilon)\|M_t^j\|_2.
\end{equation}
\end{restatable}

{\bf Proof.}
  Our goal is to bound $\|\widetilde{M}^j_t\|_2$ in terms of $\|M_t^j\|_2$.

The first moment $M_t^j$ in the original space is recursively defined as
\begin{equation*}
\small
  M_t^j \;=\; (1-\beta_1)\sum_{k=0}^{t-1}\beta_1^{\,k}\,G_{t-k}^j,
\end{equation*}
  Then, the projected first moment $\widetilde{M}^j_t$ is similarly defined as
\begin{equation*}
\small
\begin{aligned}
  \widetilde{M}^j_t \;=\; (1-\beta_1)\sum_{k=0}^{t-1}\beta_1^{\,k}\,\widetilde{G}_{t-k}^j = (1-\beta_1)\sum_{k=0}^{t-1}\beta_1^{\,k} S^\top\,G_{t-k}^j  \\
  = S^\top \!\left((1-\beta_1)\sum_{k=0}^{t-1}\beta_1^{\,k} G_{t-k}^j\right) = S^\top M_t^j
  \end{aligned}
\end{equation*}

Note that $S$ is a random matrix whose entries are i.i.d., and by Theorem~\ref{th:norm-preservation}, we know that with probability at least $1 - 2\exp(-c\epsilon^2 r)$, the following inequality holds:
\begin{equation*}
\small
  (1-\epsilon)\,\|M_t^j\|_2 \;\le\; \|S^\top\,M_t^j\|_2 \;\le\; (1+\epsilon)\,\|M_t^j\|_2.
\end{equation*}
which completes the proof. $\blacksquare$ 

\begin{restatable}[\bf Channel-Wise Second Momentum Ratio Norm Is Bounded]{lemma}{second-momentum}
\label{th:second-momentum}
For any channel \( j \) and time \( t \), if 
\(
r \geq \frac{8}{\epsilon^2} \log\left(\frac{2t}{\delta}\right),
\)
then with probability at least $1 - \delta/2$:
\begin{equation}
\small
(1-\epsilon)\|V_t^j\|_1 \leq \|\widetilde{V}_t^j\|_1 \leq (1+\epsilon)\|V_t^j\|_1
\end{equation}
\end{restatable}

{\bf Proof.}
  Channel $j$ of the second moment $V_t^j$ at iteration $t$ is defined recursively as:
\begin{equation*}
\small
  V_t^j \;=\; (1-\beta_2)\sum_{k=0}^{t-1}\beta_2^{\,k}\,\bigl(G_{t-k}^j\bigr)^{2}.
\end{equation*}
And the second momentum $\widetilde{V}_t^j$ can be defined as:
\begin{equation*}
\small
  \widetilde{V}_t^j \;=\; (1-\beta_2)\sum_{k=0}^{t-1}\beta_2^{\,k}\,\bigl(\widetilde{G}_{t-k}^j\bigr)^{2} = (1-\beta_2)\sum_{k=0}^{t-1}\beta_2^{\,k}\,\bigl(S^\top\,G_{t-k}^j\bigr)^{2}.
\end{equation*}

Computing the $\ell_1$ norm of the second moment in the original and projected space, we will have:
\begin{equation*}
\small
  \|\widetilde{V}_t^j\|_1 \;=\; \sum_{i=1}^{r}\,(1-\beta_2)\sum_{k=0}^{t-1}\beta_2^{\,k}\,\bigl(\widetilde{G}_{t-k}^{(i,j)}\bigr)^{2} = (1-\beta_2)\sum_{k=0}^{t-1}\beta_2^{\,k}\,\|\widetilde{G}_{t-k}^j\|_2^{2}.
\end{equation*}
\begin{equation*}
\small
  \|V_t^j\|_1 \;=\; \sum_{i=1}^{r}\,(1-\beta_2)\sum_{k=0}^{t-1}\beta_2^{\,k}\,\bigl(G_{t-k}^{(i,j)}\bigr)^{2} = (1-\beta_2)\sum_{k=0}^{t-1}\beta_2^{\,k}\,\|G_{t-k}^j\|_2^{2}.
\end{equation*}

By Theorem \ref{th:norm-preservation}, we know that with probablity at least $1 - 2\exp(-c\epsilon^{2}r)$:
\begin{equation*}
\small
  (1-\epsilon)\,\|G_{t-k}^j\|_2 \;\le\; \|S^\top G_{t-k}^j\|_2 \;\le\; (1+\epsilon)\,\|G_{t-k}^j\|_2,
\end{equation*}
Therefore,
\begin{equation*}
\small
  \begin{aligned}
  \|\widetilde{V}_t^j\|_1
  &= (1-\beta_2)\sum_{k=0}^{t-1}\beta_2^{\,k}\,\|\widetilde{G}_{t-k}^j\|_2^{2}
  \le (1-\beta_2)\sum_{k=0}^{t-1}\beta_2^{\,k}(1+\epsilon)^2\,\|G_{t-k}^j\|_2^{2}
  \;=\; (1+\epsilon)^2\,\|V_t^j\|_1.
  \end{aligned}
\end{equation*}
\begin{equation*}
\small
  \begin{aligned}
  \|\widetilde{V}_t^j\|_1
  &= (1-\beta_2)\sum_{k=0}^{t-1}\beta_2^{\,k}\,\|\widetilde{G}_{t-k}^j\|_2^{2}
  \ge (1-\beta_2)\sum_{k=0}^{t-1}\beta_2^{\,k}(1-\epsilon)^2\,\|G_{t-k}^j\|_2^{2}
  \;=\; (1-\epsilon)^2\,\|V_t^j\|_1.
  \end{aligned}
\end{equation*}
And thus:
\begin{equation*}
\small
  (1-\epsilon)^2\,\|V_t^j\|_1 \;\le\; \|\widetilde{V}_t^j\|_1 \;\le\; (1+\epsilon)^2\,\|V_t^j\|_1.
\end{equation*}
Now, for computing the probability across all $t$ timesteps, we know that for each $k$ the failure probability is $2\exp(-c\epsilon^{2}r)$. So, the total failure probability is
$2t\exp\!\left(-c\epsilon^{2}r\right)$.
  Setting this total failure probability to $\delta/2$ gives the condition
\begin{equation*}
\small
  r \;\ge\; \frac{8}{\epsilon^{2}}\,\log\!\frac{2t}{\delta},
\end{equation*}
  which completes the proof, similar to what we did in proof of Theorem \ref{th:distance-preservations}. $\blacksquare$

  Now that we have stated these two lemmas, we get back to the proof of Theorem.
  \bounded*
  {\bf Proof.} 
  Given the definition of $\phi$ in \eqref{eq:phi}:
  \begin{equation*}
  \small
      \frac{\phi_t(\widetilde{G}_t^j)}{\phi_t(G_t^j)} = \frac{{\|\widetilde G^{j\,O}_{t}\|}/{\|\widetilde G^j_{t}\|}}{{\| G^{j\,O}_{t}\|}/{\| G^j_{t}\|}} = \frac{\|\widetilde G^{j\,O}_{t}\|}{\| G^{j\,O}_{t}\|} \cdot \frac{\| G^j_{t}\|}{\|\widetilde G^j_{t}\|}
  \end{equation*}

  Applying Theorem~\ref{th:norm-preservation}:
\begin{equation}
\label{bound1}
\small
  \frac{\|G_t^j\|}{\|{\widetilde{G}}_t^j\|} \;\in\; \left[\frac{1}{1+\epsilon},\;\frac{1}{1-\epsilon}\right].
\end{equation}

Replacing the optimizer outputs by Adam-like update:
\begin{equation*}
    \frac{\|\widetilde G^{j\,O}_{t}\|^2_2}{\| G^{j\,O}_{t}\|^2_2} = \frac{\|(\widetilde{M}_t^j/\sqrt{\widetilde{V}_t^j})\|_2^2}{\|(M_t^j/\sqrt{V_t^j})\|^2_2} = 
    \frac{\sum_{i=1}^{r}\,(\widetilde M^{i, j}_t)^2 / \widetilde V^{i, j}_t}
    {\sum_{i=1}^{n}\,(M_t^{i, j})^2   / V_t^{i,j}}
\end{equation*}

  % \subparagraph{Case 1: SGD with Momentum.}
  % The variance term is non-existent and \eqref{eq:second-factor} simplifies to
  % \[
  % \frac{\|\tilde{R}_t[:,j]\|_2^{\,2}}{\|\tilde{G}_t[:,j]\|_2^{\,2}}
  % \;=\;
  % \frac{\|M^R_t[:,j]\|_2^{\,2}}{\|M_t[:,j]\|_2^{\,2}}.
  % \]
  % Applying Theorem~\ref{thm:first-moment} (first-moment preservation),
  % \[
  % 1-\epsilon \;\le\; \frac{\|M^R_t[:,j]\|}{\|M_t[:,j]\|} \;\le\; 1+\epsilon,
  % \]
  % so the second factor obeys
  % \[
  % \frac{\|\tilde{R}_t[:,j]\|}{\|\tilde{G}_t[:,j]\|} \;\in\; [\,1-\epsilon,\;1+\epsilon\,].
  % \]
Now, because this ration involves element-wise division, we will use a trick, that was used in prior work \cite{zhu2025apollosgdlikememoryadamwlevel}. 
Based on observations in previous works, including 
Adam-mini \citep{zhang2024adamminiusefewerlearning} and GaLore-mini \citep{huang2024galoremini}, we know that variance can be effectively approximated channel-wise. So, we replace each element, by an average, calculated based on the $\ell_1$ norm of $V_t$ on each channel:
\begin{equation*}
    \small
    \frac{\|\widetilde G^{j\,O}_{t}\|^2_2}{\| G^{j\,O}_{t}\|^2_2} \approx 
    \frac{\sum_{i=1}^{r}\,(\widetilde M^{i, j}_t)^2 / (\|\widetilde V^{j}_t\|_1/r)}
    {\sum_{i=1}^{n}\,(M_t^{i, j})^2   / (\|V_t^{j}\|_1/m)}
\end{equation*}
By this approximation, we can now factorize the constant values, and then, we can rewrite our estimation as:
\begin{equation*}
    \small
    \frac{\|\widetilde G^{j\,O}_{t}\|^2_2}{\| G^{j\,O}_{t}\|^2_2} \approx \frac{r}{m} \cdot
    \frac{\|V_t^{j}\|_1}{\|\widetilde V^{j}_t\|_1} \cdot
    \frac{\|\widetilde{M_t^j}\|_2^2}{\|M_t^j\|_2^2} 
\end{equation*}

Then, by applying the bounds from Lemmas \ref{th:first-momentum} and \ref{th:second-momentum}, we can derive:
\begin{equation*}
    \small
    \begin{aligned}
         \frac{\|V_t^{j}\|_1}{\|\widetilde V^{j}_t\|_1} \in \left[\frac{1}{(1+\varepsilon)^2}, \frac{1}{(1-\varepsilon)^2}\right], \\
         \frac{\|\widetilde{M_t^j}\|_2^2}{\|M_t^j\|_2^2} \in \left[ (1-\varepsilon)^2, (1+\varepsilon)^2 \right]
    \end{aligned}
\end{equation*}
then, 
\begin{equation*}
\small
    \frac{m}{r} \cdot \frac{\|\widetilde G^{j\,O}_{t}\|^2_2}{\| G^{j\,O}_{t}\|^2_2} \in 
    \left[ 
    \frac{(1-\varepsilon)^2}{(1+\varepsilon)^2}, 
    \frac{(1+\varepsilon)^2}{(1-\varepsilon)^2}
    \right]
\end{equation*}
and:
\begin{equation}
\label{bound2}
\small
    \sqrt{\frac{m}{r}} \cdot \frac{\|\widetilde G^{j\,O}_{t}\|_2}{\| G^{j\,O}_{t}\|_2} \in 
    \left[ 
    \frac{(1-\varepsilon)}{(1+\varepsilon)}, 
    \frac{(1+\varepsilon)}{(1-\varepsilon)}
    \right]
\end{equation}
Finally, the bounding of the requested term can be achieved by combining the bounds derived in \eqref{bound1} and \eqref{bound2}, and the proof is complete. $\blacksquare$

Regarding the probability of success, this bound is derived from Lemmas \ref{bound1} and \ref{bound2}. Following the union bound argument presented in Lemma \ref{bound2}, one can verify that under the specified condition for r, the theorem holds with probability at least $1-\delta$.

  % \begin{remark}
  % The bound contains the constant factor $n/r$, which suggests that we should scale the gradient to maintain consistent behavior with AdamW under structured learning-rate updates. This gradient scaling factor can
  %  be folded into the learning rate. When $r$ is too small compared to $n$ — as in our APOLLO-Mini case, which uses a rank-$1$ space — we explicitly assign the scaling factor (e.g., $128$).
  % \end{remark}

  % \begin{remark}
  % The requirement that $r$ grows sublinearly as $\log(t)$ ensures that even for large $t$ the rank $r$ does not grow excessively. Empirically, however, we find our
  % method is not sensitive to rank selection; even a rank of $256$ is sufficient to train LLaMA-7B with 150k steps. This can be explained by recent
  % Adam-mini~\cite{zhang2024adammini}, which observes that the variance need not be precise and a block-wise approximation suffices, indicating that the variance
  % approximation error can be tolerated well.
  % \end{remark}

  % \begin{remark}
  % Here, we assume the projection matrix is fixed over time step $t$. GaLore~\cite{zhao2024galore} also derives their theorem with the same assumption. However, as
  % acknowledged in GaLore, using the same projection matrix for the entire training may limit the directions in which the weights can grow. Therefore, empirically, as
  % in GaLore, we periodically resample $P$ over $T$ iterations to introduce new directions. Unlike GaLore, which uses time-consuming SVD-based updates, we can simply
  % re-sample $P$ from the Gaussian distribution by changing the random seed.
  % \end{remark}

%% file: Images/gradient_energy_256/tex.tex
\begin{figure}[t]
 % First Rown
 \scriptsize
    \centering
    \begin{subfigure}{0.24\textwidth}
        \centering
        \includegraphics[width=\linewidth]{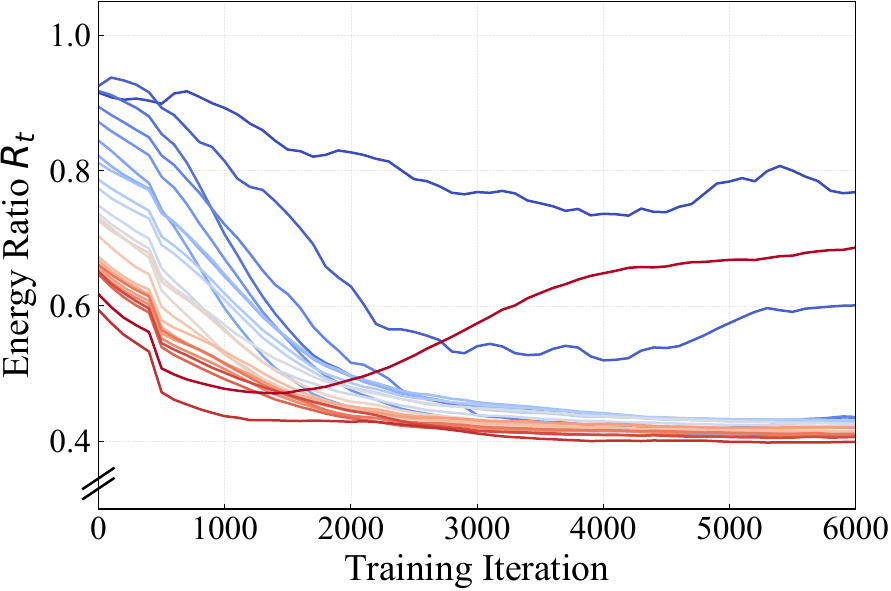}
        \caption{\scriptsize Attention-Output Proj.}
    \end{subfigure}
    \begin{subfigure}{0.24\textwidth}
        \centering
        \includegraphics[width=\linewidth]{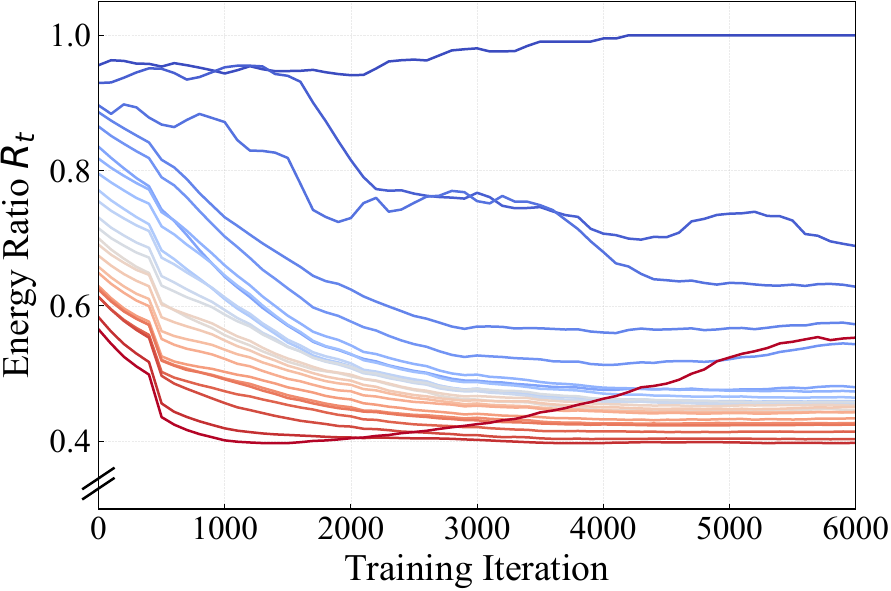}
        \caption{\scriptsize Attention-Value Proj.}
    \end{subfigure}
    \begin{subfigure}{0.24\textwidth}
        \centering
        \includegraphics[width=\linewidth]{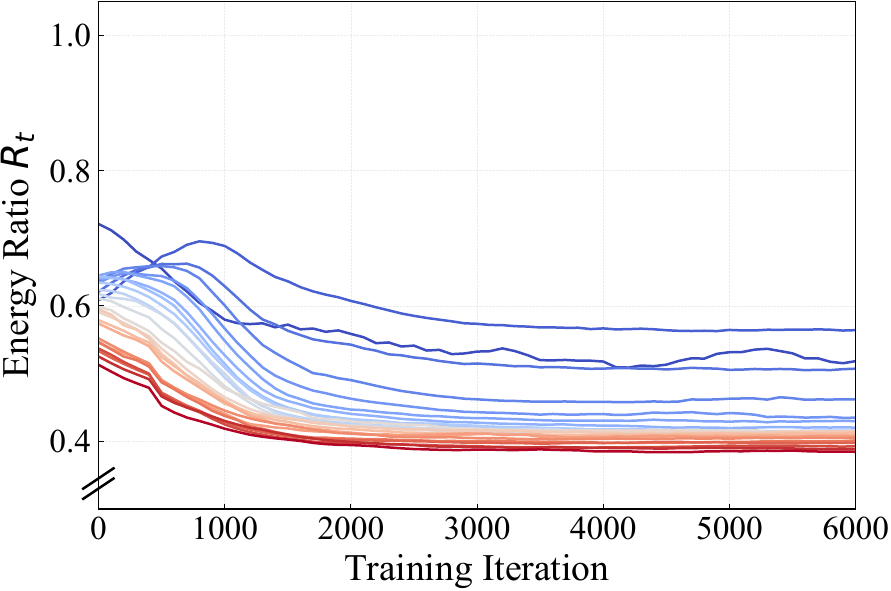}
        \caption{\scriptsize Attention-Query Proj.}
    \end{subfigure}
    \begin{subfigure}{0.24\textwidth}
        \centering
        \includegraphics[width=\linewidth]{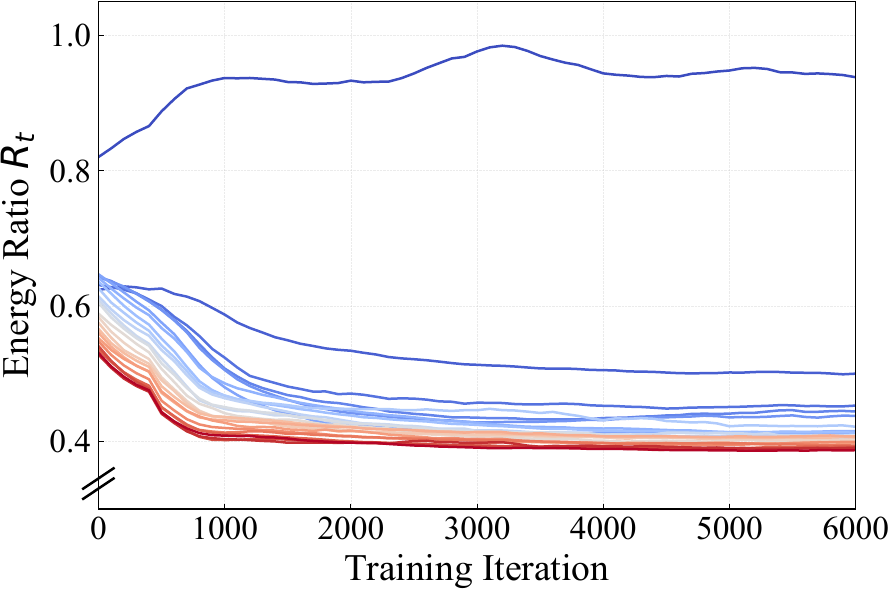}
        \caption{\scriptsize Attention-Key Proj.}
    \end{subfigure}

    % Second row
    \begin{subfigure}{0.24\textwidth}
        \centering
        \includegraphics[width=\linewidth]{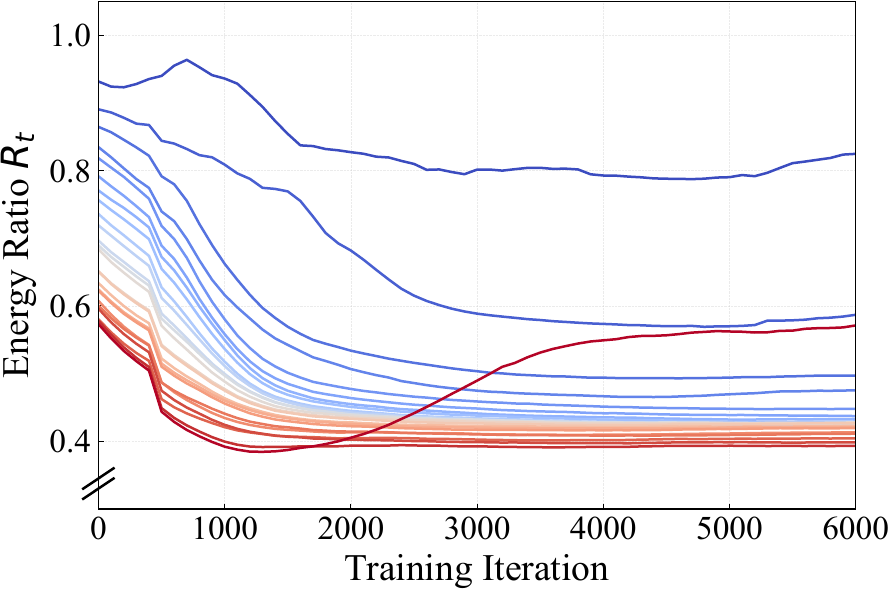}
        \caption{\scriptsize MLP-Gate Proj.}
    \end{subfigure}
    \begin{subfigure}{0.24\textwidth}
        \centering
        \includegraphics[width=\linewidth]{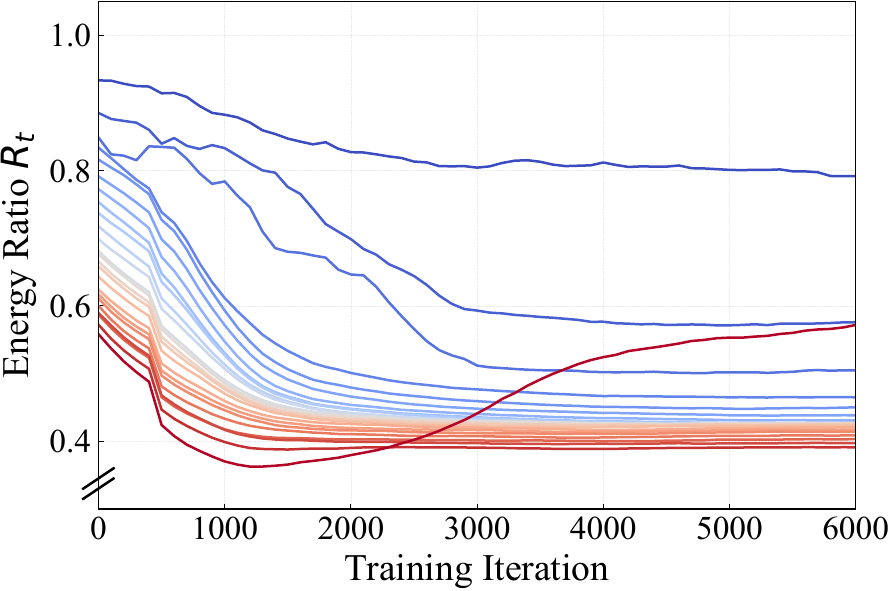}
        \caption{\scriptsize MLP-Up Proj.}
    \end{subfigure}
    \begin{subfigure}{0.24\textwidth}
        \centering
        \includegraphics[width=\linewidth]{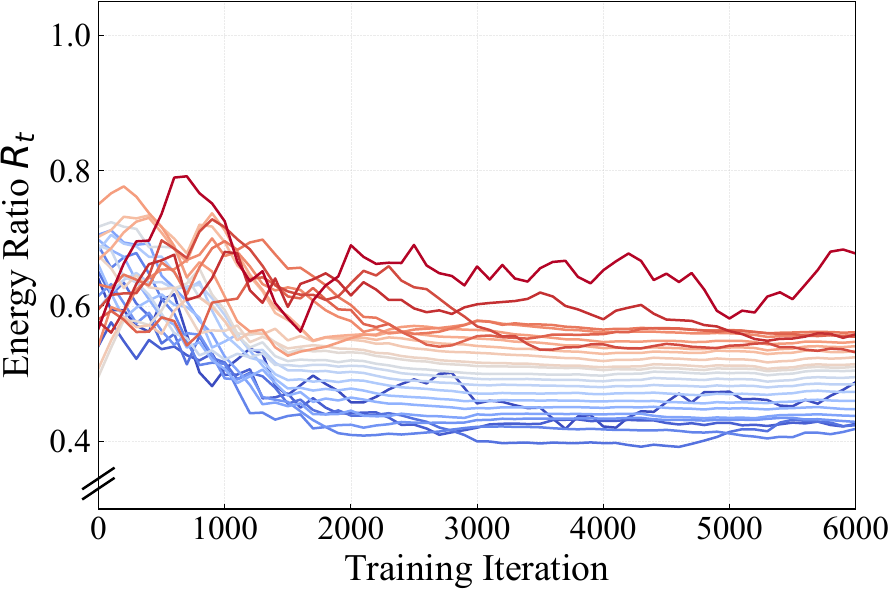}
        \caption{\scriptsize MLP-Down Proj.}
    \end{subfigure}
    \begin{subfigure}{0.24\textwidth}
        \centering
        \includegraphics[width=\linewidth]{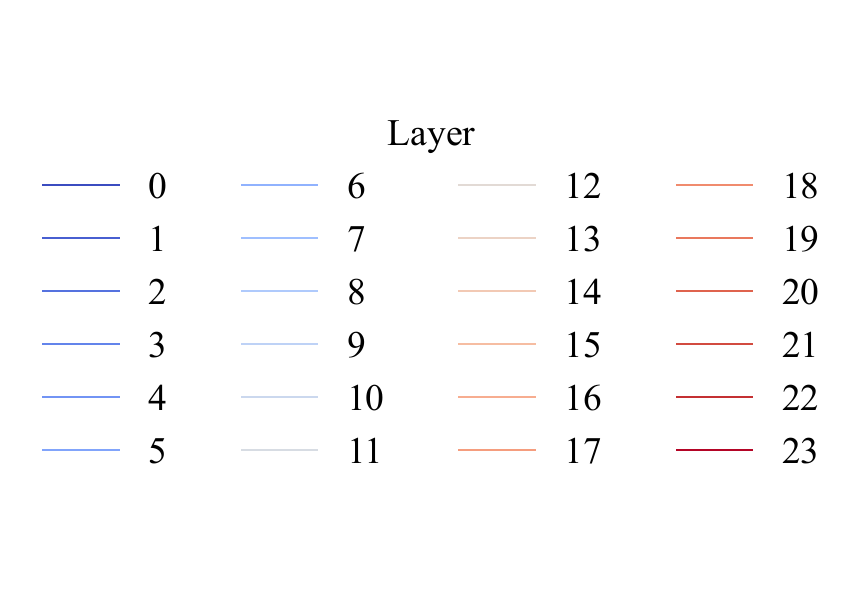}
        \caption{\scriptsize Legend}
    \end{subfigure}

    \caption{Each decoder layer stack includes seven layer types in the Llama-1B model. The plots show the fraction of gradient-matrix energy explained by a {rank 256} approximation. Despite a high lower bound for this rank, this fraction declines over training, and deeper layers generally exhibit smaller fractions.}
    \label{fig:energy_fraction-256}
\end{figure}

%% file: Images/gradient_energy_1024/tex.tex
\begin{figure}[t]
 % First Rown
 \scriptsize
    \centering
    \begin{subfigure}{0.24\textwidth}
        \centering
        \includegraphics[width=\linewidth]{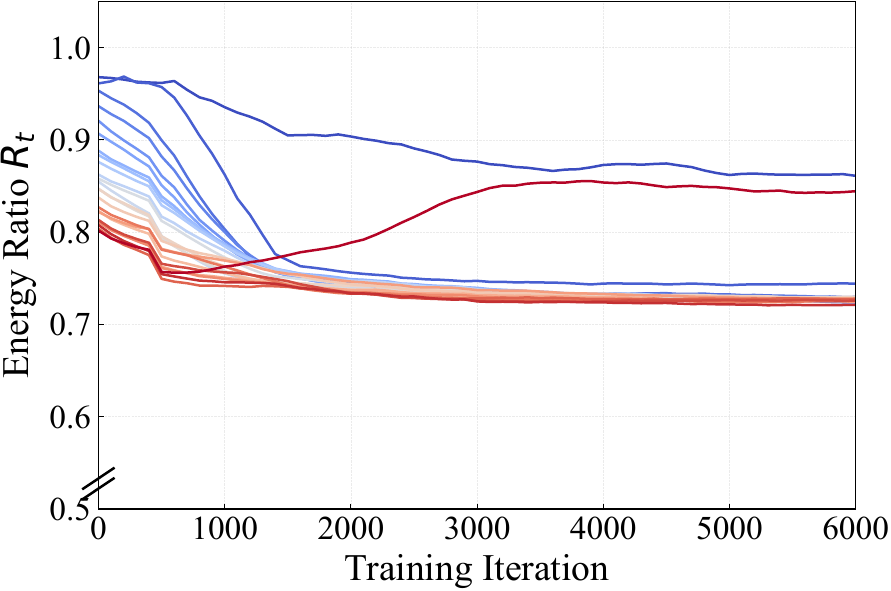}
        \caption{\scriptsize Attention-Output Proj.}
    \end{subfigure}
    \begin{subfigure}{0.24\textwidth}
        \centering
        \includegraphics[width=\linewidth]{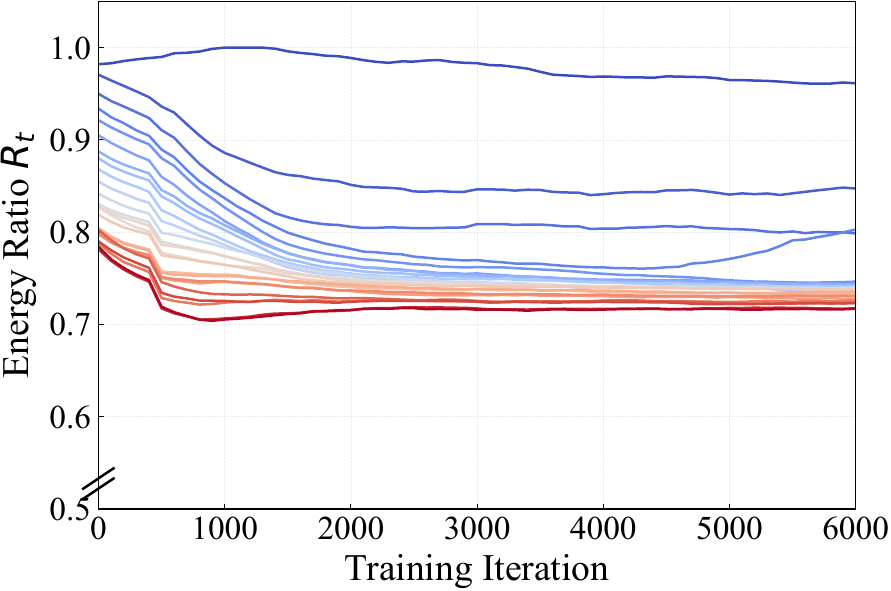}
        \caption{\scriptsize Attention-Value Proj.}
    \end{subfigure}
    \begin{subfigure}{0.24\textwidth}
        \centering
        \includegraphics[width=\linewidth]{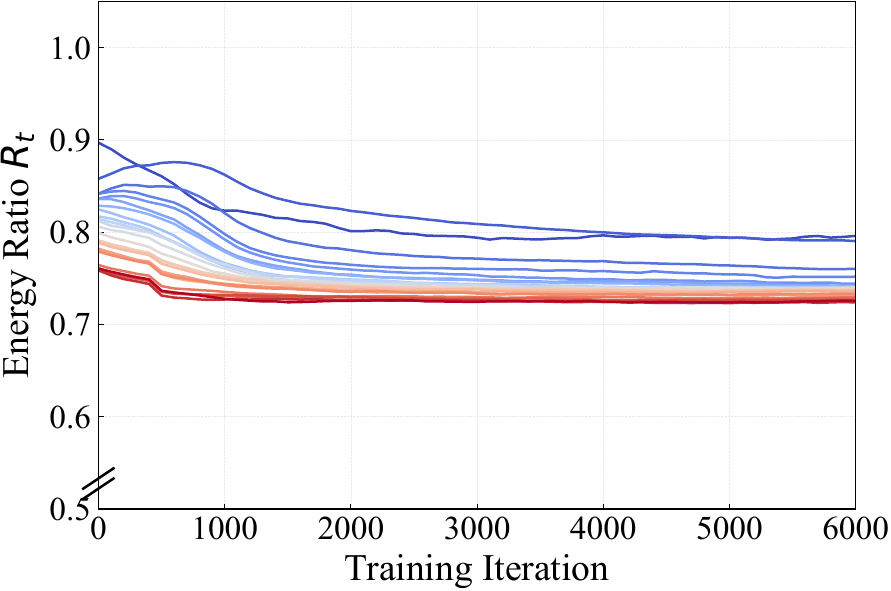}
        \caption{\scriptsize Attention-Query Proj.}
    \end{subfigure}
    \begin{subfigure}{0.24\textwidth}
        \centering
        \includegraphics[width=\linewidth]{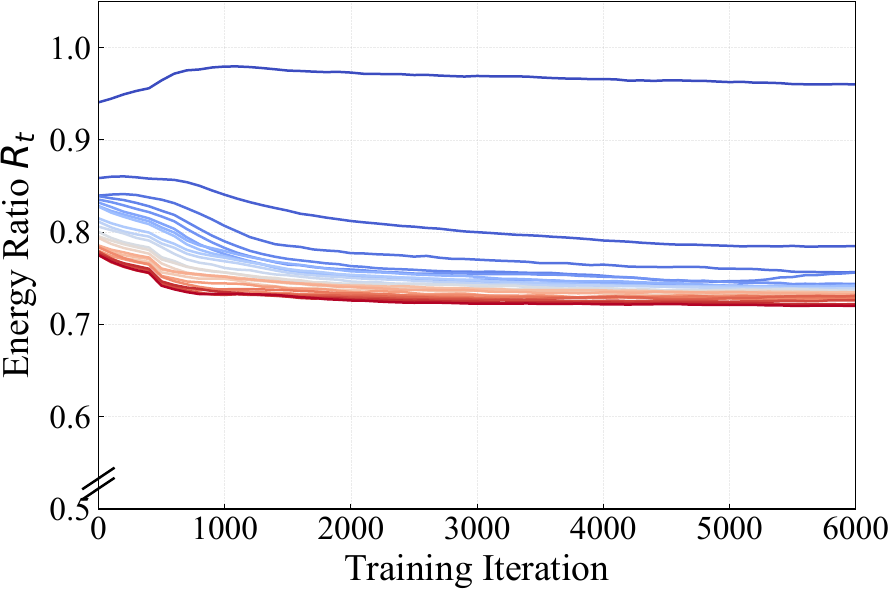}
        \caption{\scriptsize Attention-Key Proj.}
    \end{subfigure}

    % Second row
    \begin{subfigure}{0.24\textwidth}
        \centering
        \includegraphics[width=\linewidth]{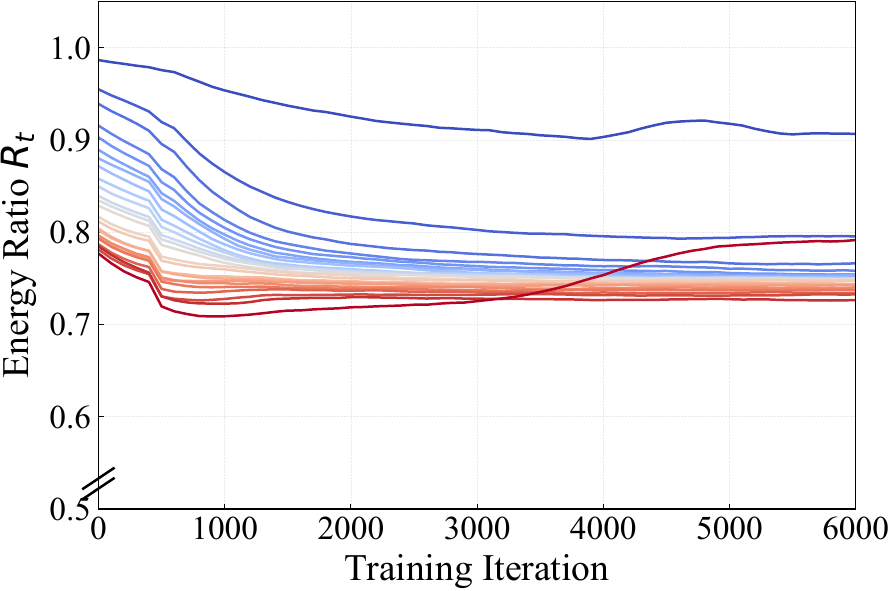}
        \caption{\scriptsize MLP-Gate Proj.}
    \end{subfigure}
    \begin{subfigure}{0.24\textwidth}
        \centering
        \includegraphics[width=\linewidth]{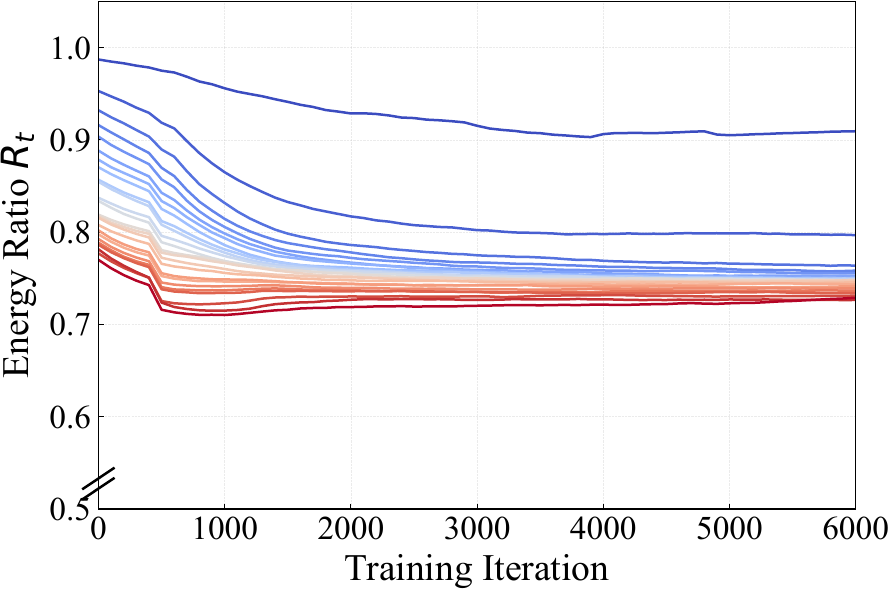}
        \caption{\scriptsize MLP-Up Proj.}
    \end{subfigure}
    \begin{subfigure}{0.24\textwidth}
        \centering
        \includegraphics[width=\linewidth]{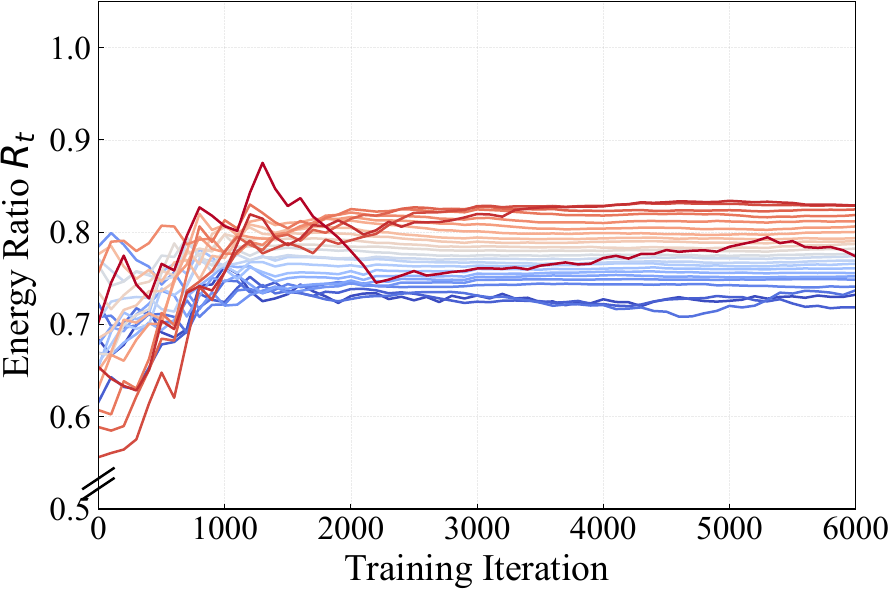}
        \caption{\scriptsize MLP-Down Proj.}
    \end{subfigure}
    \begin{subfigure}{0.24\textwidth}
        \centering
        \includegraphics[width=\linewidth]{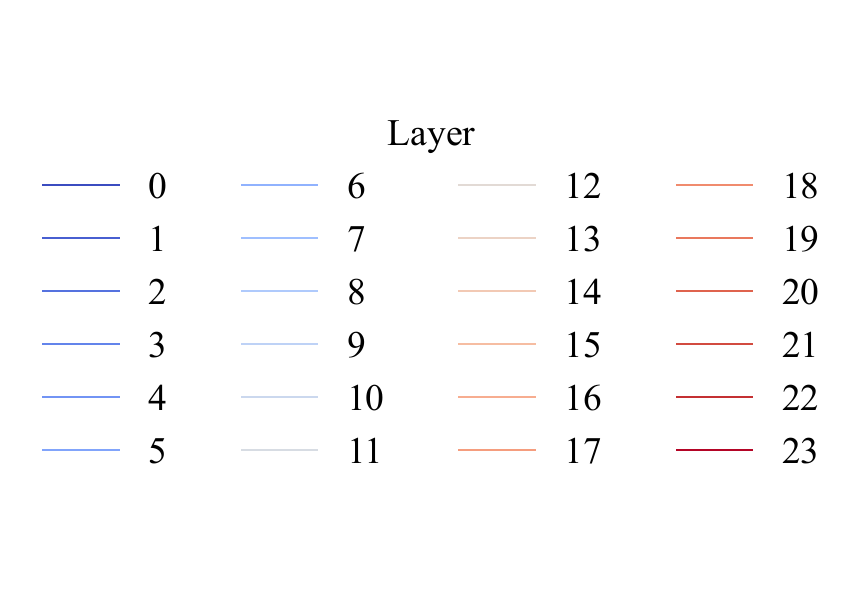}
        \caption{\scriptsize Legend}
    \end{subfigure}

    \caption{Each decoder layer stack includes seven layer types in the Llama-1B model. The plots show the fraction of gradient-matrix energy explained by a {rank 1024} approximation. Despite large rank for this model size and a high lower bound, this fraction declines over training, and deeper layers generally exhibit smaller fractions.}
    \label{fig:energy_fraction-1024}
\end{figure}

%% file: Tablels/r_abl.tex
\begin{table}[!h]
\centering
\caption{Final evaluation loss (\(\downarrow\)) after pre-training a Llama-1B architecture for 10k iterations and with different subspace ranks.}
\label{tab:r-abl}
\begin{tabular}{lccc}
\toprule
\textbf{Method} & \textbf{r = 256} & \textbf{r = 512} & \textbf{r = 1024} \\
\midrule
GrassWalk   
    &  4.62  & 4.52  & 4.43 \\
GrassJump
    & 4.54   & 4.50  & 4.45 \\
\bottomrule
\end{tabular}
\vspace{-10pt}
\end{table}

%% file: Images/norm_growth_limiter/tex.tex
\begin{figure}[t]
 % First Rown
 \small
    \centering
    \begin{subfigure}{0.49\textwidth}
    \label{fig:1b-ngl}
        \centering
        \includegraphics[width=\linewidth]{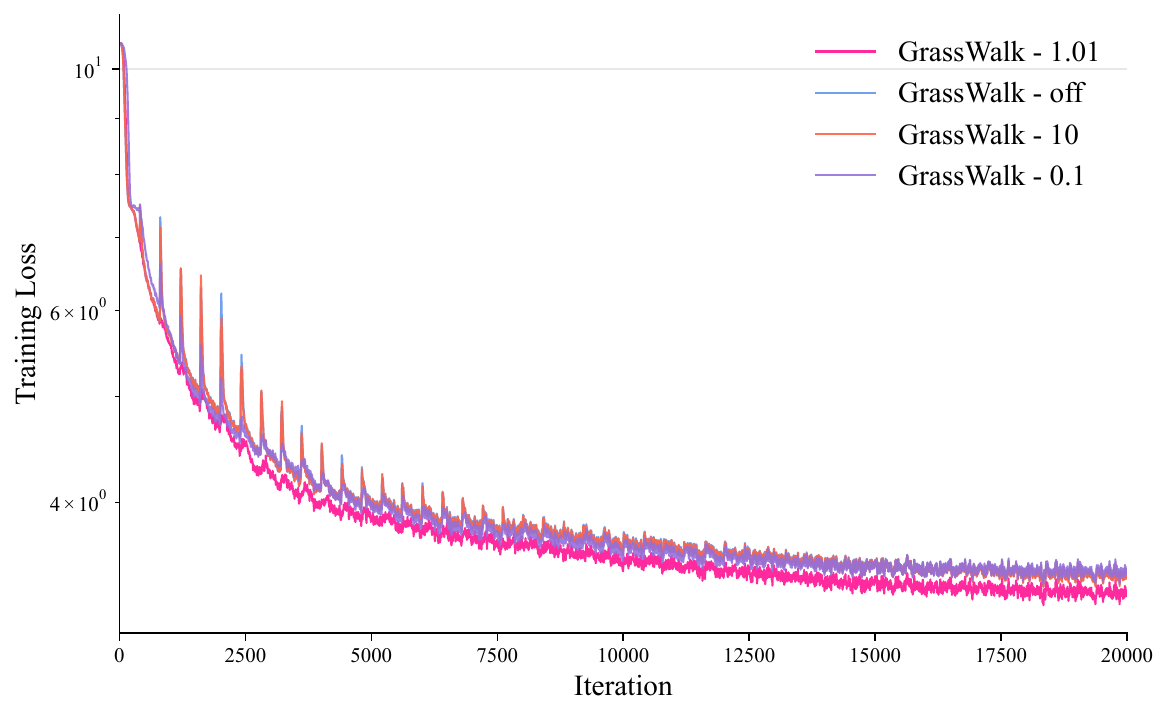}
        \caption{\scriptsize GrassWalk}
    \end{subfigure}
    \begin{subfigure}{0.49\textwidth}
    \label{fig:7b-ngl}
        \centering
        \includegraphics[width=\linewidth]{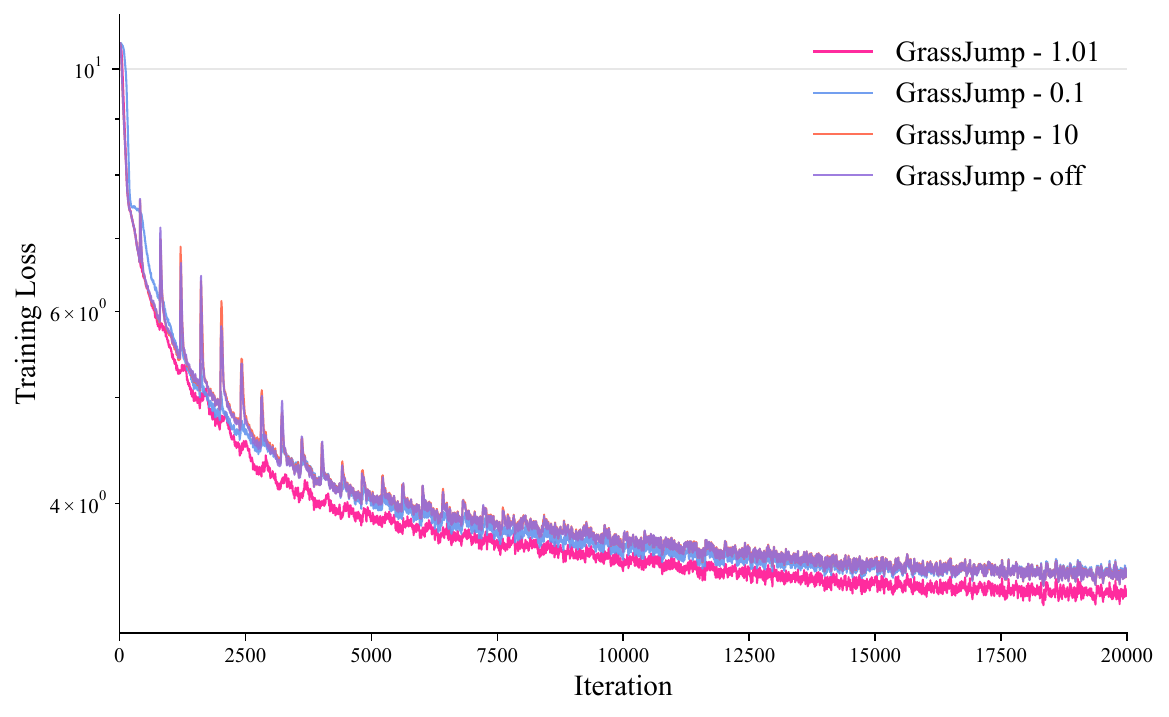}
        \caption{\scriptsize GrassJump}
    \end{subfigure}
    \caption{Training dynamics of pre-training a Llama-1B architecture with rank 512 for 10k iterations using different values for norm-growth limiter $\zeta$.}
    \label{fig:zeta}
\end{figure}

%% file: Tablels/z_abl.tex
\begin{table}[!h]
\centering
\caption{Final evaluation loss (\(\downarrow\)) after pre-training a Llama-1B architecture for 10k iterations with $r = 512$ and with different values for the norm-growth limiter $\zeta$.}
\label{tab:z-abl}
% \resizebox{\linewidth}{!}{
\begin{tabular}{lcccc}
\toprule
\textbf{Method} & \textbf{$\zeta = 0.1$} & \textbf{$\zeta = 1.01$} & \textbf{$\zeta = 10$} & \textbf{$\zeta = \infty$} \\
\midrule
GrassWalk  
    &  3.44 & 3.26 & 3.41 & 3.42 \\
GrassJump 
    &  3.45 & 3.30 & 3.44 & 3.44 \\
\bottomrule
\end{tabular}
% }
% \vspace{-10pt}
\end{table}

%% file: Tablels/e_abl.tex
\begin{table}[!h]
\centering
\caption{Final evaluation loss (\(\downarrow\)) after pre-training a Llama-1B architecture for 10k iterations with $r = 512$ using GrassWalk with diffent subspace update step-size $\eta$.}
\label{tab:e-abl}
% \resizebox{\linewidth}{!}{
\begin{tabular}{lcccc}
\toprule
\textbf{Method} & \textbf{$\eta = 10$} & \textbf{$\eta = 100$} & \textbf{$\eta = 1000$} & \textbf{$\eta = 10000$} \\
\midrule
GrassWalk  
    &  3.9812 & 3.9786 & 3.9815 & 3.9890 \\
\bottomrule
\end{tabular}
% }
\vspace{-10pt}
\end{table}

%% file: Images/long_run_pretraining/tex.tex
\begin{figure}[t]
 % First Rown
 \small
    \centering
    \begin{subfigure}{0.49\textwidth}
    \label{fig:1b-long}
        \centering
        \includegraphics[width=\linewidth]{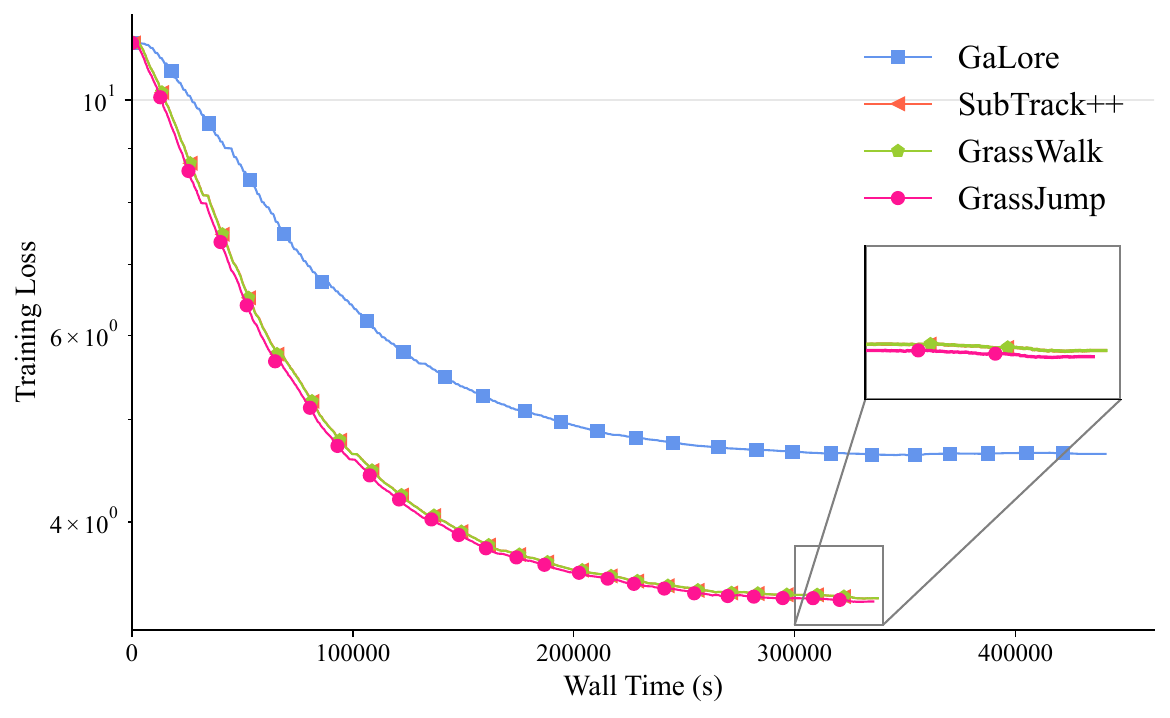}
        \caption{\scriptsize Training Loss vs. Wall-Time}
    \end{subfigure}
    \begin{subfigure}{0.49\textwidth}
    \label{fig:7b-long}
        \centering
        \includegraphics[width=\linewidth]{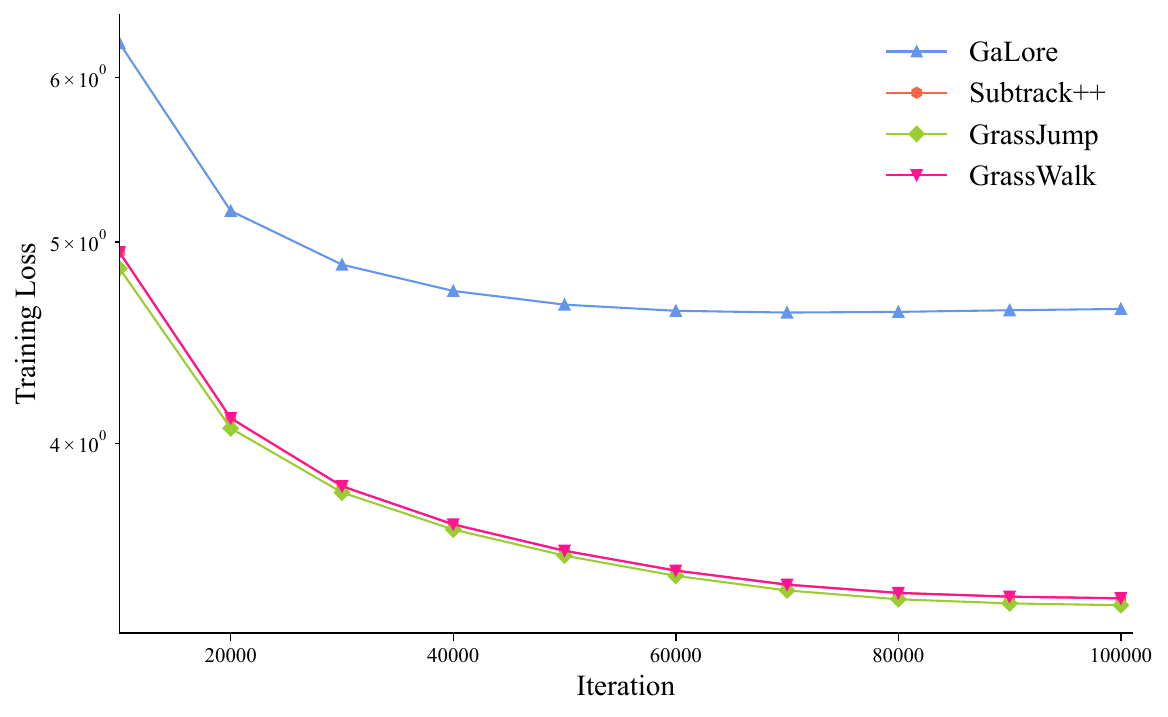}
        \caption{\scriptsize Evaluation Loss vs. Training Iterations}
    \end{subfigure}
    \caption{Training and evaluation dynamics of pre-training a Llama-7B architecture with rank 1024 for 100k iterations across different methods.}
    \label{fig:long-run}
\end{figure}

%% file: Tablels/hyperparameters.tex
\begin{table*}[!h]
    \caption{Hyperparameters of pre-training Llama-based and Qwen architectures.}
    \small
    \centering
    \label{tab:pt_hyperparameters}
    \begin{tabular}{l|c|ccc}
    \midrule
    \midrule
    & & Llama-1B   & Llama-7B & Qwen-1.5B   \\
    \midrule
    \midrule
    Architectural & Hidden         &2048 &4096 &1536 \\
    Parameters &Intermediate   &5461 &11008 & 8960 \\
    &Heads         &24 &32 &12 \\
    &Layers        &32 &32 &28 \\
    \midrule
    Shared Parameters & Learning Rate &1e-4 &1e-4 &1e-4   \\
    & Batch Size    &32 &8 &16   \\
    & Gradient Accumulation & 2 &4 &4 \\
    & Iterations & 10k & 100k & 10k \\
    & Gradient Clipping & \multicolumn{3}{c}{1.0} \\
    & Warmup Steps & \multicolumn{3}{c}{1000} \\
    & scale & \multicolumn{3}{c}{0.25} \\
    & dtype & \multicolumn{3}{c}{bfloat16} \\
    \midrule
    Low-Rank Optimizer &Rank          &512  &1024  &256   \\
    Methods Parameters & Subspace Update Interval & 100 & 500 & 100 \\
    & \methodname Step-Size & \multicolumn{3}{c}{10000} \\
    \bottomrule
    \end{tabular}
\end{table*}

%% file: Images/tangent_fro/tex.tex
\begin{figure}[t]
 % First Rown
 \scriptsize
    \centering
    \begin{subfigure}{0.24\textwidth}
        \centering
        \includegraphics[width=\linewidth]{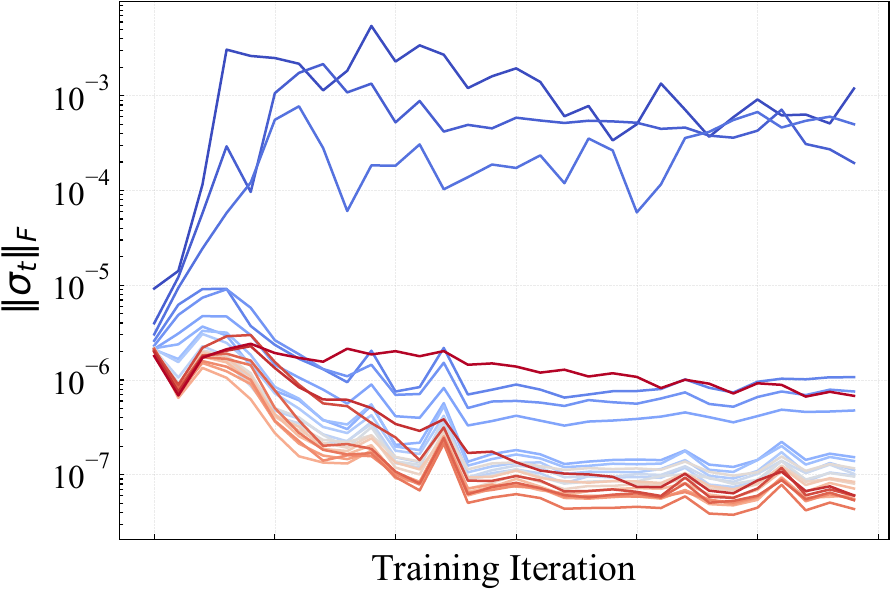}
        \caption{\scriptsize Attention-Output Proj.}
    \end{subfigure}
    \begin{subfigure}{0.24\textwidth}
        \centering
        \includegraphics[width=\linewidth]{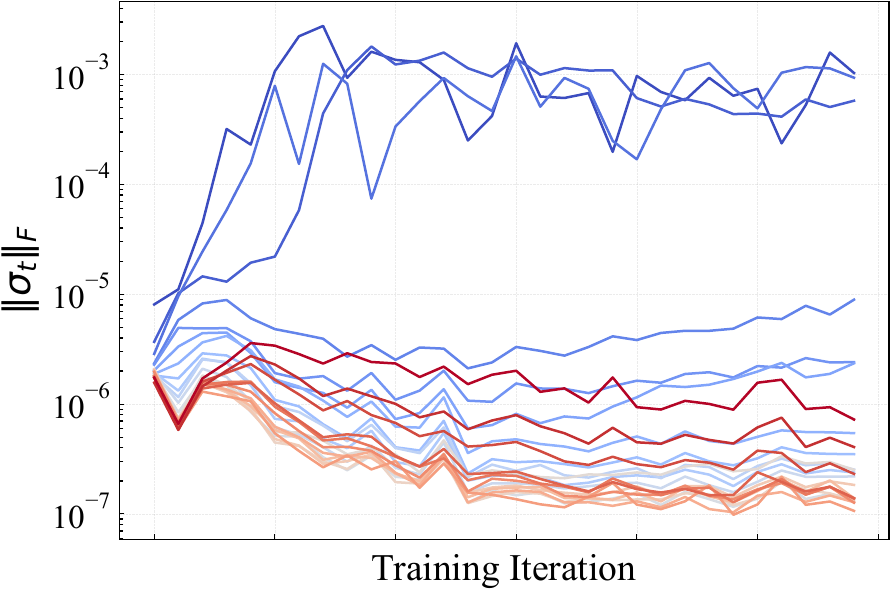}
        \caption{\scriptsize Attention-Value Proj.}
    \end{subfigure}
    \begin{subfigure}{0.24\textwidth}
        \centering
        \includegraphics[width=\linewidth]{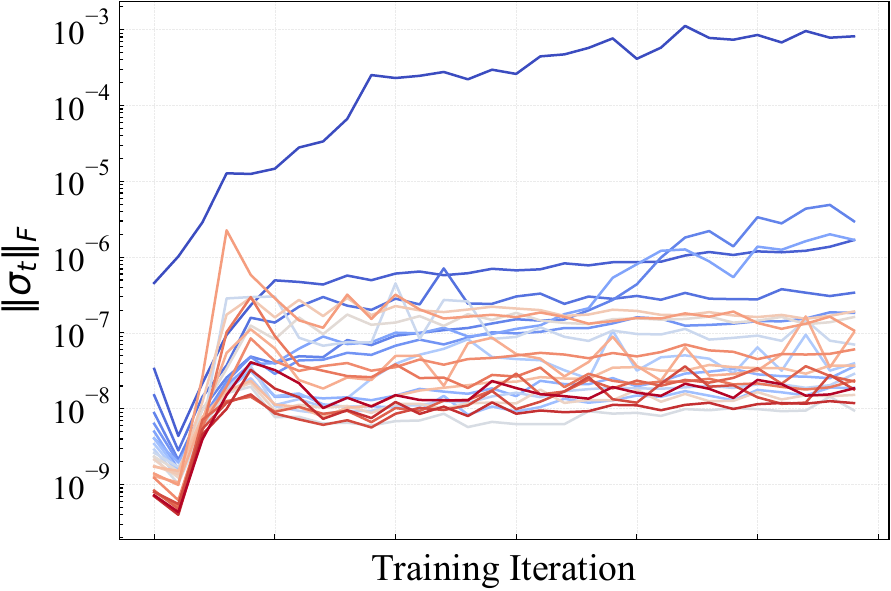}
        \caption{\scriptsize Attention-Query Proj.}
    \end{subfigure}
    \begin{subfigure}{0.24\textwidth}
        \centering
        \includegraphics[width=\linewidth]{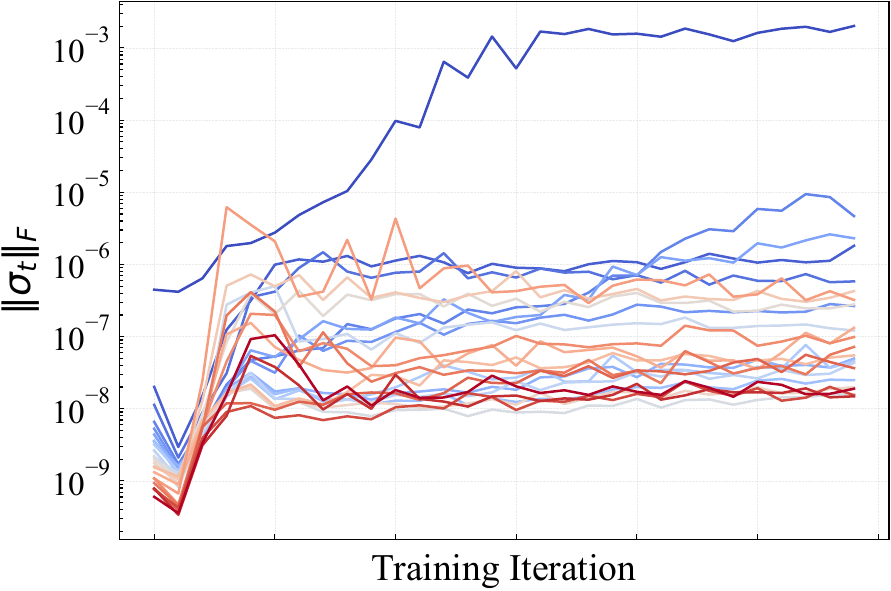}
        \caption{\scriptsize Attention-Key Proj.}
    \end{subfigure}

    % Second row
    \begin{subfigure}{0.24\textwidth}
        \centering
        \includegraphics[width=\linewidth]{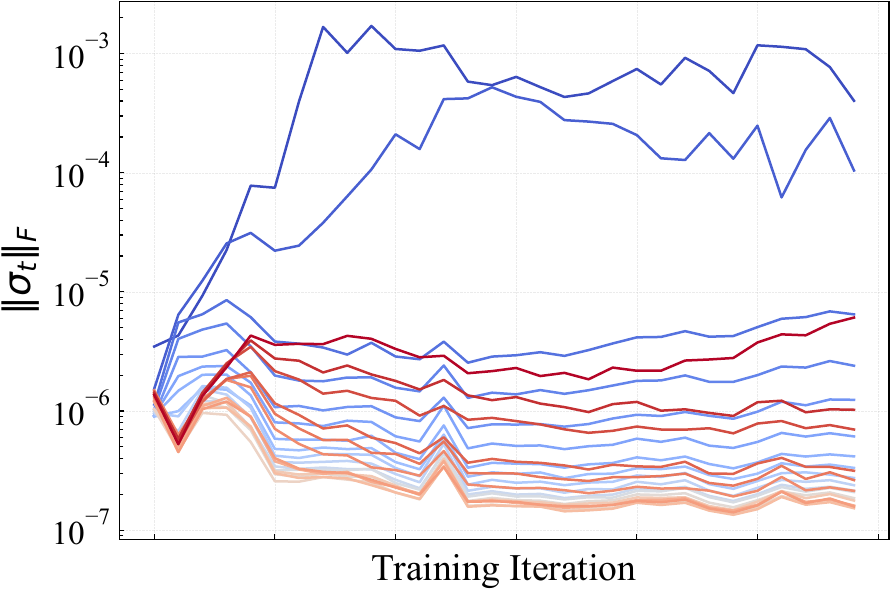}
        \caption{\scriptsize MLP-Gate Proj.}
    \end{subfigure}
    \begin{subfigure}{0.24\textwidth}
        \centering
        \includegraphics[width=\linewidth]{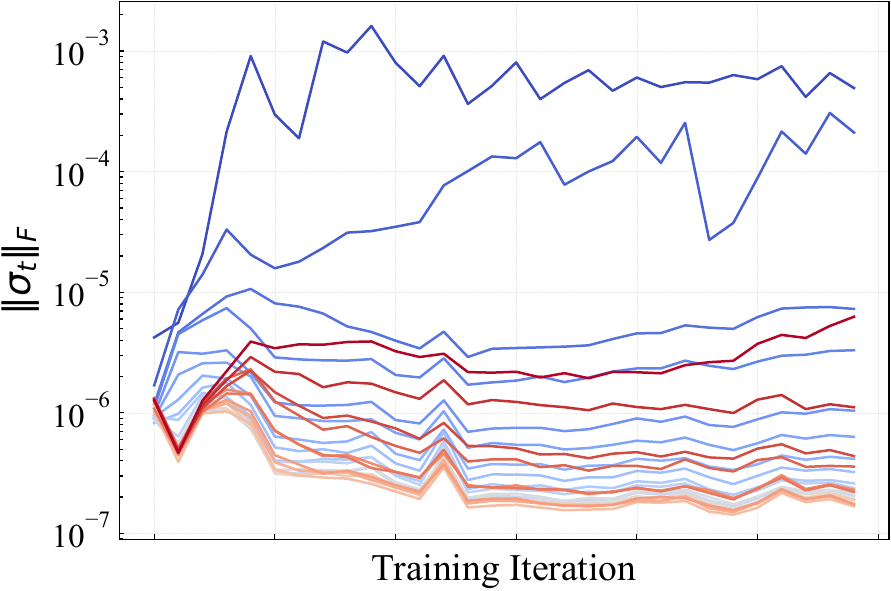}
        \caption{\scriptsize MLP-Up Proj.}
    \end{subfigure}
    \begin{subfigure}{0.24\textwidth}
        \centering
        \includegraphics[width=\linewidth]{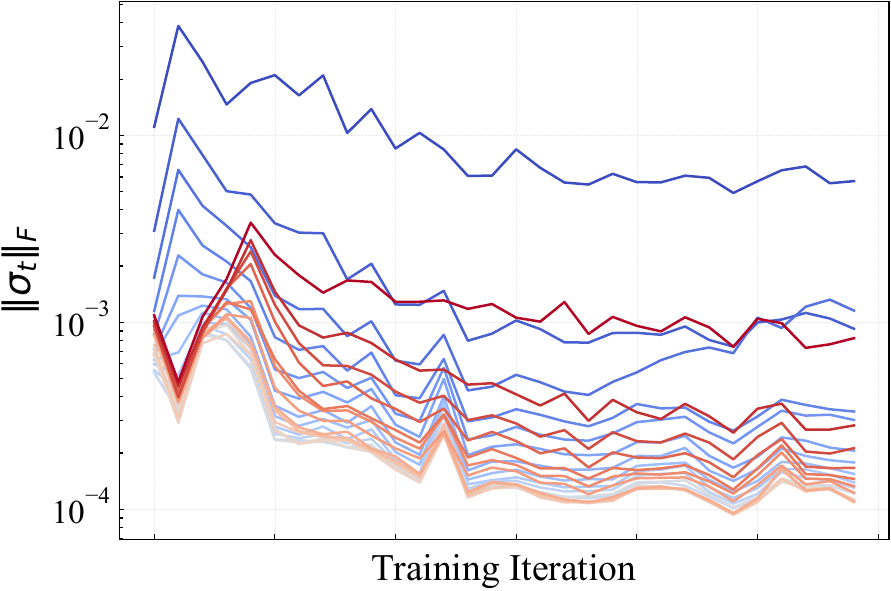}
        \caption{\scriptsize MLP-Down Proj.}
    \end{subfigure}
    \begin{subfigure}{0.24\textwidth}
        \centering
        \includegraphics[width=\linewidth]{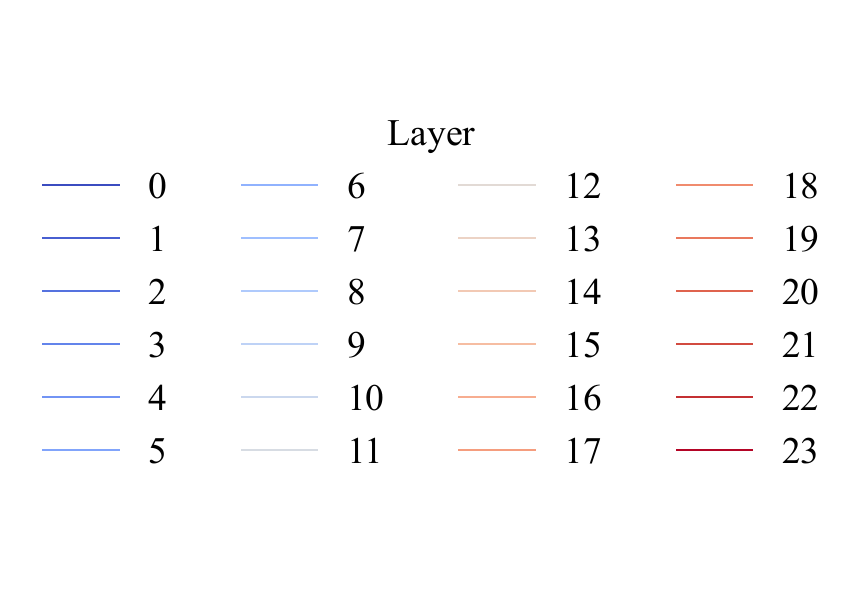}
        \caption{\scriptsize Legend}
    \end{subfigure}

    \caption{Frobenius norm of singular values of the subspace estimation error derivative across different projection layers in the LLaMA-1B architecture. Each plot shows the norm within a given layer type (aggregated across all 24 decoder layers) as training progresses. Almost all the layers demonstrate extremely small values, indicating the flat curvature of the gradient subspace optimization space.}
    \label{fig:tangent-fro}
\end{figure}

%% file: checklist.tex
\section*{NeurIPS Paper Checklist}

\begin{enumerate}

\item {\bf Claims}
    \item[] Question: Do the main claims made in the abstract and introduction accurately reflect the paper's contributions and scope?
    \item[] Answer: \answerYes{} % Replace by \answerYes{}, \answerNo{}, or \answerNA{}.
    \item[] Justification: We included the overview of method, achieved results, and the scope of experiments in abstract and introduction.

\item {\bf Limitations}
    \item[] Question: Does the paper discuss the limitations of the work performed by the authors?
    \item[] Answer: \answerYes{} % Replace by \answerYes{}, \answerNo{}, or \answerNA{}.
    \item[] Justification: The limitations and future directions is included in the last section.

\item {\bf Theory assumptions and proofs}
    \item[] Question: For each theoretical result, does the paper provide the full set of assumptions and a complete (and correct) proof?
    \item[] Answer: \answerYes{} % Replace by \answerYes{}, \answerNo{}, or \answerNA{}.
    \item[] Justification: The theoretical proofs are aligned with prior works while improving over them, and the process of derivation are explained in detail.

    \item {\bf Experimental result reproducibility}
    \item[] Question: Does the paper fully disclose all the information needed to reproduce the main experimental results of the paper to the extent that it affects the main claims and/or conclusions of the paper (regardless of whether the code and data are provided or not)?
    \item[] Answer: \answerYes{} % Replace by \answerYes{}, \answerNo{}, or \answerNA{}.
    \item[] Justification: All resources and hyperparameters used for testing the proposed method and baselines, and also the analysis are provided in the appendix.

\item {\bf Open access to data and code}
    \item[] Question: Does the paper provide open access to the data and code, with sufficient instructions to faithfully reproduce the main experimental results, as described in supplemental material?
    \item[] Answer: \answerNo{} % Replace by \answerYes{}, \answerNo{}, or \answerNA{}.
    \item[] Justification: We will release the code upon acceptance.

\item {\bf Experimental setting/details}
    \item[] Question: Does the paper specify all the training and test details (e.g., data splits, hyperparameters, how they were chosen, type of optimizer) necessary to understand the results?
    \item[] Answer: \answerYes{}{} % Replace by \answerYes{}, \answerNo{}, or \answerNA{}.
    \item[] Justification: All the details regarding resources and hyperparameters for the proposed method and baselines are included in the appendix.

\item {\bf Experiment statistical significance}
    \item[] Question: Does the paper report error bars suitably and correctly defined or other appropriate information about the statistical significance of the experiments?
    \item[] Answer: \answerNo{} % Replace by \answerYes{}, \answerNo{}, or \answerNA{}.
    \item[] Justification: As the experiments of these paper are highly resource- and time-consuming, performing statistical test was not possible for this version. However, a variety of experiments are included to ensure generalization.

\item {\bf Experiments compute resources}
    \item[] Question: For each experiment, does the paper provide sufficient information on the computer resources (type of compute workers, memory, time of execution) needed to reproduce the experiments?
    \item[] Answer: \answerYes{} % Replace by \answerYes{}, \answerNo{}, or \answerNA{}.
    \item[] Justification: All the necessary information is concluded in appendix.
    
\item {\bf Code of ethics}
    \item[] Question: Does the research conducted in the paper conform, in every respect, with the NeurIPS Code of Ethics \url{https://neurips.cc/public/EthicsGuidelines}?
    \item[] Answer: \answerYes{} % Replace by \answerYes{}, \answerNo{}, or \answerNA{}.
    \item[] Justification: Yes, it totally conform.

\item {\bf Broader impacts}
    \item[] Question: Does the paper discuss both potential positive societal impacts and negative societal impacts of the work performed?
    \item[] Answer: \answerNA{} % Replace by \answerYes{}, \answerNo{}, or \answerNA{}.
    \item[] Justification: This is a foundational research, and further investigation for preserving safety is left for future work.
    
\item {\bf Safeguards}
    \item[] Question: Does the paper describe safeguards that have been put in place for responsible release of data or models that have a high risk for misuse (e.g., pre-trained language models, image generators, or scraped datasets)?
    \item[] Answer: \answerNA{} % Replace by \answerYes{}, \answerNo{}, or \answerNA{}.
    \item[] Justification: It is a foundational research and do not poses such risks.

\item {\bf Licenses for existing assets}
    \item[] Question: Are the creators or original owners of assets (e.g., code, data, models), used in the paper, properly credited and are the license and terms of use explicitly mentioned and properly respected?
    \item[] Answer: \answerYes{} % Replace by \answerYes{}, \answerNo{}, or \answerNA{}.
    \item[] Justification: We have cited properly. 

\item {\bf New assets}
    \item[] Question: Are new assets introduced in the paper well documented and is the documentation provided alongside the assets?
    \item[] Answer: \answerNA{} % Replace by \answerYes{}, \answerNo{}, or \answerNA{}.
    \item[] Justification: No assest is provided at this step.

\item {\bf Crowdsourcing and research with human subjects}
    \item[] Question: For crowdsourcing experiments and research with human subjects, does the paper include the full text of instructions given to participants and screenshots, if applicable, as well as details about compensation (if any)? 
    \item[] Answer: \answerNA{} % Replace by \answerYes{}, \answerNo{}, or \answerNA{}.
    \item[] Justification: This term does not apply to our research. 

\item {\bf Institutional review board (IRB) approvals or equivalent for research with human subjects}
    \item[] Question: Does the paper describe potential risks incurred by study participants, whether such risks were disclosed to the subjects, and whether Institutional Review Board (IRB) approvals (or an equivalent approval/review based on the requirements of your country or institution) were obtained?
    \item[] Answer: \answerNA{} % Replace by \answerYes{}, \answerNo{}, or \answerNA{}.
    \item[] Justification: This term does not apply to our research. 

\item {\bf Declaration of LLM usage}
    \item[] Question: Does the paper describe the usage of LLMs if it is an important, original, or non-standard component of the core methods in this research? Note that if the LLM is used only for writing, editing, or formatting purposes and does \emph{not} impact the core methodology, scientific rigor, or originality of the research, declaration is not required.
    %this research? 
    \item[] Answer: \answerNA{} % Replace by \answerYes{}, \answerNo{}, or \answerNA{}.
    \item[] Justification: This term does not apply to our research. 

\end{enumerate}